\documentclass[sigconf]{acmart}
\usepackage{amsmath}
\usepackage{url}
\usepackage{marvosym}
\usepackage{graphicx}
\usepackage{xcolor}
\usepackage{multirow}
\usepackage{tabularx}
\usepackage{algorithm}
\usepackage{algorithmic}
\usepackage{subcaption}
\usepackage{amsthm}
\usepackage{enumitem}

\newcolumntype{C}{>{\centering\arraybackslash}X} % centered version of 'X' col. type

\usepackage{thmtools}
\usepackage{thm-restate}
\theoremstyle{plain}
\newtheorem{problem}{Problem}
\newtheorem{theorem}{Theorem}[section]

\newtheorem{lemma}{Lemma}[section]

\newtheorem{definition}{Definition}[section]

\AtBeginDocument{%
  \providecommand\BibTeX{{%
    \normalfont B\kern-0.5em{\scshape i\kern-0.25em b}\kern-0.8em\TeX}}}

\copyrightyear{2025}
\acmYear{2025}\setcopyright{acmlicensed}\acmConference[KDD '25]{Proceedings of the 31st ACM
SIGKDD Conference on Knowledge Discovery and Data Mining V.1}{August 3--7,
2025}{Toronto, ON, Canada}
\acmBooktitle{Proceedings of the 31st ACM SIGKDD Conference on Knowledge
Discovery and Data Mining V.1 (KDD '25), August 3--7, 2025, Toronto, ON, Canada}
\acmDOI{10.1145/3690624.3709289}
\acmISBN{979-8-4007-1245-6/25/08}

\begin{document}

\title{Stealing Training Graphs from Graph Neural Networks}


\author{Minhua Lin}
\email{minhualin@psu.edu}
\affiliation{
\institution{The Pennsylvania State University}
\city{State College}
\country{USA}}

\author{Enyan Dai}
\email{enyandai@hkust-gz.edu.cn}
\affiliation{
\institution{Hong Kong University of Science and Technology (Guangzhou)}
\city{Guangzhou}
\country{China}}
% \authornote{Both authors contribute equally to this paper.}

\author{Junjie Xu}
\email{junjiexu@psu.edu}
\affiliation{
\institution{The Pennsylvania State University}
\city{State College}
\country{USA}}

\author{Jinyuan Jia}
\email{jinyuan@psu.edu}
\affiliation{
\institution{The Pennsylvania State University}
\city{State College}
\country{USA}}

\author{Xiang Zhang}
\email{xzz89@psu.edu}
\affiliation{
\institution{The Pennsylvania State University}
\city{State College}
\country{USA}}

\author{Suhang Wang}
\email{szw494@psu.edu}
\affiliation{
    \institution{The Pennsylvania State University}
    \city{State College}
    \country{USA}
}

% \author{Minhua Lin{\textsuperscript{\textdagger}}, Enyan Dai{$^{\ddagger}$}, Junjie Xu{\textsuperscript{\textdagger}}, Jinyuan Jia{\textsuperscript{\textdagger}}, Xiang Zhang{\textsuperscript{\textdagger}}, Suhang Wang{\textsuperscript{\textdagger}}}
% % \authornote{Both authors contributed equally to this research.}
% \email{{minhualin,junjiexu,jinyuan,xzz89,szw494}@psu.edu, enyandai@hkust-gz.edu.cn}
% \affiliation{%
% {\textsuperscript{\textdagger}}
%   \institution{The Pennsylvania State University, USA}
%   % \streetaddress{P.O. Box 1212}
%   % \city{State College}
%   % \state{Ohio}
%   \country{}
%   % \\
%   {$^{\ddagger}$}
%   \institution{Hong Kong University of Science and Technology (Guangzhou), China}
%   % \city{Guangzhou}
%   \country{}
%   % \postcode{43017-6221}
\renewcommand{\shortauthors}{Minhua Lin, Enyan Dai, Junjie Xu, Jinyuan Jia, Xiang Zhang, and Suhang Wang.}

% \begin{abstract}
%   A clear and well-documented \LaTeX\ document is presented as an
%   article formatted for publication by ACM in a conference proceedings
%   or journal publication. Based on the ``acmart'' document class, this
%   article presents and explains many of the common variations, as well
%   as many of the formatting elements an author may use in the
%   preparation of the documentation of their work.
% \end{abstract}
\begin{abstract}
Graph Neural Networks (GNNs) have shown promising results in modeling graphs in various tasks. The training of GNNs, especially on specialized tasks such as bioinformatics, demands extensive expert annotations, which are expensive and usually contain sensitive information of data providers. The trained GNN models are often shared for deployment in the real world. As neural networks can memorize the training samples, the model parameters of GNNs have a high risk of leaking private training data. Our theoretical analysis shows the strong connections between trained GNN parameters and the training graphs used, confirming the training graph leakage issue. However, explorations into training data leakage from trained GNNs are rather limited. 
Therefore, we investigate a novel problem of stealing graphs from trained GNNs. To obtain high-quality graphs that resemble the target training set, a graph diffusion model with diffusion noise optimization is deployed as a graph generator. Furthermore, we propose a selection method that effectively leverages GNN model parameters to identify training graphs from samples generated by the graph diffusion model. Extensive experiments on real-world datasets demonstrate the effectiveness of the proposed framework in stealing training graphs from the trained GNN. The code is publicly available at \url{https://github.com/ventr1c/GraphSteal}.
\end{abstract}









\maketitle

\section{Introduction}
Graph-structured data pervade numerous real-world applications such as social networks~\cite{hamilton2017inductive}, finance systems~\cite{wang2019semi}, and molecular graphs~\cite{wang2022molecular}.
Graph Neural Networks (GNNs) have shown promising results in modeling graphs by adopting a message passing scheme~\cite{kipf2016semi,xu2018powerful,wang2024efficient,liang2024survey}, which updates a node's representation by aggregating information from its neighbors. The learned representation can preserve both node attributes and local graph structural information, facilitating various tasks, such as node classification~\cite{} and graph classifications~\cite{errica2019fair}. 

\begin{figure}
    \centering
    \includegraphics[width=0.86\linewidth]{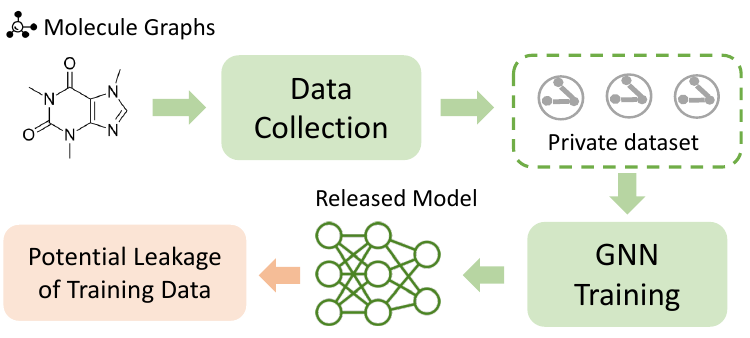}
    \vskip -1em
    \caption{Illustration of GNNs training and releasing.}
    \vskip -1em
    \label{fig:illustration_of_GNN_training_releasing}
\end{figure}

As shown in Fig~\ref{fig:illustration_of_GNN_training_releasing}, generally, abundant training data is required to train a high-performance GNN model. This becomes particularly expensive for critical applications such as molecule property prediction, which demands expert annotations. Moreover, the training data may hold sensitive information belonging to its providers. Consequently, protecting the privacy of the training data is imperative.  After training on the private training data, the trained GNN model is often released for downstream applications. For example, a well-trained molecule property predictor may be made open source, supporting the direct deployment for the designed task or model initialization for other tasks. However, neural networks can memorize the training data, even when they have great generalization ability~\cite{zhang2021understanding}. It is also demonstrated that the trained parameters of MLP are linear combinations of the derivatives of the network at a set of training data points~\cite{Lyu2020Gradient}, causing potential private data leakage. Our theoretical analysis in Theorem~\ref{thm:relationship_parameter_training_data} further shows that the above observations can be extended to GNNs. Hence, releasing GNN models potentially threatens the privacy of the private training data. Therefore, in this paper, we investigate a novel problem of stealing training graphs from trained GNN models when the model architecture and parameters are known/released.

Several initial efforts are made in model inversion attacks, which aim to reconstruct graph topologies~\cite{zhang2022inference} or infer the sensitive node attributes attributes~\cite{duddu2020quantifying}. However, they are proposed to reconstruct the train graph/node from node/graph embeddings, which cannot be applied for training graph stealing as embeddings of private training data are unavailable for attackers. Recently, GraphMI~\cite{zhang2021graphmi} proposes to reconstruct the adjacency matrix of the training graph in a white-box setting where the trained model parameters are available. However, GraphMI requires all the node attributes of the training graph to be available for topology reconstruction. This assumption is impractical in real-world applications. Therefore, in this paper, we propose to steal the training graphs from the trained GNN model without any information of the training data, which is not explored by existing works.

However, it is non-trivial to perform the graph stealing attack. There are two major challenges to be addressed. 
\textit{First},  it is challenging to ensure both the realism and quality of graphs reconstructed from the trained GNN model.
Existing model inversion attacks on GNNs mainly focus at the node level, concentrating on reconstructing either links~\cite{he2021stealing,zhang2021graphmi} or node attributes~\cite{duddu2020quantifying,zhang2022inference} of the target graphs. However, this focus is inadequate for graph stealing attacks because graph-structured data comprises both graph topology and node attributes, rendering these methods insufficient for comprehensive graph-level reconstruction.
Our empirical analysis in Sec.~\ref{sec:comparisions_with_baselines} further demonstrates that current methods encounter difficulties in providing realistic and high-quality graphs during graph-level reconstruction. 
\textit{Second},
how to effectively utilize the trained GNN model parameters to extract the training graph information? Previous works on model inversion attacks on GNNs~\cite{duddu2020quantifying,he2021stealing,zhang2021graphmi} typically rely on comprehensive side information of target samples for graph reconstruction. The investigations on extracting training data information from trained GNN parameters without any partial training data information poses a new challenge.

In an effort to address the aforementioned challenges, we propose a novel \textit{Graph} \textit{Steal}ing attack framework (GraphSteal). Specifically, to overcome the challenge of reconstructing high-quality graphs, we employ a graph diffusion model as the graph generator. In addition, this graph diffusion model used in GraphSteal adopts a diffusion noise optimization algorithm, which can produce a set of candidate graphs that more closely resemble the training set of the target GNN.
Moreover, according to our theoretical analysis in Theorem~\ref{thm:relationship_parameter_training_data}, there is a strong connection between the GNN model parameters and training data. Based on the theorem, we propose a model parameter-guided graph selection method, which leverages the parameters of GNNs to identify the training graphs from the candidate graph set generated by the graph diffusion model.
In summary, our main contributions are:
\vspace{-1em}
% \vskip -5em
\begin{itemize}[leftmargin=*]
    \item We study a new problem of stealing training graphs from a trained GNN without any partial information on training data;
    \item We propose GraphSteal, a novel attack framework that can effectively recreate high-quality graphs that are part of the target training dataset by leveraging parameters of GNN models; and
    \item Extensive experiments on various real-world datasets demonstrate the effectiveness of our proposed GraphSteal in accurately reconstructing private training graphs of trained GNN models.
\end{itemize}

\section{Related Work}
\textbf{Graph Neural Networks.} Graph Neural Networks (GNNs)~\cite{kipf2016semi,ying2018graph,liu2022spherical, javaloy2023learnable, lee2024transitivity,xu2024llm,lin2024certifiably} have shown great power in modeling graph-structured data, which have been deployed to various applications such as social network analysis~\cite{fan2019graph,dai2021say,zhang2024robustness,lin2024trojan,liang2024hawkes}, drug discovery~\cite{xu2023graph,dai2021towards,dai2022towards,dai2023unnoticeable,zhang2024rethinking} and energy network analysis~\cite{dai2022graph}. The success of GNNs lies in the message-passing mechanism, which iteratively aggregates a node's neighborhood information to refine the node's representations. For example, GCN~\cite{kipf2016semi} combines a node's neighborhood information by averaging their representations with the target center nodes. To improve the expressivity of GNNs, GIN~\cite{xu2018powerful} further incorporates a hidden layer in combining the neighbors' information.  
Inspired by the success of transformers in modeling image and text~\cite{devlin2018bert,dosovitskiy2020image,lin2024decoding,fang2023annotations,fang2025your}, graph transformer is also proposed~\cite{muller2023attending,rampavsek2022recipe,zhang2020graph,ying2021transformers,fang2024not,liang2024simple}. Generally, graph transformer has a global attention mechanism for graph embedding learning. It shows promising results, especially in molecule property prediction.  Despite the great achievements, GNNs could be vulnerable to privacy attacks, which largely constrain their adoption in safety-critical domains such as bioinformatics and financial analysis. 

\vspace*{0.3em}

\noindent\textbf{Privacy Attacks on GNNs.} 
% \suhang{also cite more recent papers} 
The training of GNNs requires a large amount of data. In critical domains such as bioinformatics~\cite{li2021braingnn} and healthcare~\cite{li2020graph}, sensitive data of users will be collected to train a powerful GNN to facilitate the services. However, recent studies show that privacy attacks can extract various private information from GNN models~\cite{dai2022comprehensive,dai2024pregip,olatunji2021membership,duddu2020quantifying,dai2023unified,dai2022learning,hou2024adversarial,liang2024mgksite}, threatening the privacy of users. For example, membership inference attacks~\cite{olatunji2021membership,duddu2020quantifying,he2021node, wu2024link} can identify whether a target sample is in the training set. This is achieved by learning a binary classifier on patterns such as posteriors that training and test samples exhibit different distributions. Model extract attacks~\cite{wu2022model} aim to steal the target model by building a model that behaves similarly to the target model. Model inversion attacks, also known as reconstruction attacks, try to infer the private information of the test/training data. For example, ~\citeauthor{duddu2020quantifying}~\cite{duddu2020quantifying} propose to infer sensitive attributes of users from their node embeddings. The reconstruction of the adjacency matrix from node embeddings is studied in~\cite{zhang2022inference}. GraphMI~\cite{zhang2021graphmi} further considers a white-box setting for adjacency matrix reconstruction, where the target GNN model parameters and node features are available for attackers. Our GraphSteal lies in the model inversion attack. However, GraphSteal is inherently different from the aforementioned methods because (\textit{\bf i}) we focus on a new problem of stealing training graphs from the trained GNN without even partial information of training set; (\textit{\bf ii}) we propose a novel framework GraphSteal which designs a diffusion noise optimization algorithm and a model parameter-guided graph selection mechanism to reconstruct high-quality training graphs of the target GNN model.

\vspace*{0.3em}
% \subsection{Graph Diffusion Models}
\noindent \textbf{Graph Diffusion Models.} Diffusion models have showcased their exceptional performance on various tasks, including image generation~\cite{ho2020denoising, song2021scorebased}, text-to-image generation~\cite{ramesh2022hierarchical, balaji2022ediffi} and video generation~\cite{ho2022imagen}. 
Generally, a diffusion model has two phases: (i) the diffusion phase, where noise is incrementally added to the clean data; and (ii) the denoising phase, where a model learns to predict and remove the noise. After trained, the denoise module can effectively generate realistic samples with noises as input. 
Recent works try to extend diffusion models to graphs. For example, \citet{vignac2023digress} proposes DiGress to bridge the gap between the discreteness of graph structure and the continuity of the diffusion and denoising process. \citet{jo2022score} uses a system of stochastic differential equations to model the joint distribution of nodes and edges. \citet{luo2023fast} conduct the graph diffusion process from the spectral domain. EDM~\cite{hoogeboom2022equivariant} and GeoDiff~\cite{xu2022geodiff} aim to preserve the equivariant information of 3D graphs.

\section{Backgrounds and Preliminaries}
\noindent\textbf{Notations}. Let $\mathcal{G}=(\mathcal{V},\mathcal{E}, \mathbf{X})$ denote an attributed graph, where $\mathcal{V}=\{v_1,\dots,v_{n}\}$ is the set of $n$ nodes, $\mathcal{E} \subseteq \mathcal{V} \times \mathcal{V}$ is the set of edges, and $\mathbf{X}=\{\mathbf{x}_1,...,\mathbf{x}_N\}$ is the set of node attributes with $\mathbf{x}_i$ being the node attribute of $v_i$. $\mathbf{A} \in \mathbb{R}^{n \times n}$ is the adjacency matrix of $\mathcal{G}$, where $\mathbf{A}_{ij}=1$ if nodes ${v}_i$ and ${v}_j$ are connected; otherwise $\mathbf{A}_{ij}=0$. In this paper, we focus on graph stealing attack from a GNN classifier trained on a private labeled graph dataset $\mathcal{D}_L = \{\mathcal{G}_1,\dots,\mathcal{G}_l\}$  with  $\mathcal{Y}_{L}=\{y_1,\dots,y_l\}$ being the corresponding labels. $y_i \in \{1,2,\ldots,C\}$ is the label of $\mathcal{G}_i$, where $C$ is the number of classes.

\subsection{Graph Diffusion Model}
\label{sec:graph_diffusion_model}
Graph diffusion models have demonstrated a robust capability in generating realistic graphs. In this paper, we mainly employ DiGress~\cite{vignac2023digress}, a popular discrete denoising graph diffusion model, as the graph generator.  The diffusion process of DiGress is a Markov process consisting of successive
graphs edits (edge addition or deletion, node or edge category edit) that can occur independently on each node or edge. As the nodes and edges are considered to belong to one of the given categories, the noises are modeled as transition matrices $(Q^1, \ldots, Q^T )$, where $\left[{Q}^t\right]_{i j}$ is the probability of jumping from category $i$ to category $j$.
To invert this diffusion process, it trains a graph transformer network to predict the clean graph from a noisy input. 
Specifically, DiGress diffuses each node and edge feature separately by applying transition matrices. The diffusion process at $t$-th step can treated as sampling node type and edge type of $\mathcal{G}^t$ from the categorical distributions $\mathbf{X}^{t-1} Q_\mathbf{X}^t$ and $\mathcal{E}^{t-1} {Q}_\mathcal{E}^t$,  respectively, which can be written as
\begin{equation}
\label{eq:diffusion_one_step_forward_process}
q(\mathcal{G}^t \mid \mathcal{G}^{t-1})=(\mathbf{X}^{t-1} Q_\mathbf{X}^t, \mathcal{E}^{t-1} {Q}_\mathcal{E}^t),
\end{equation}
where $\mathcal{G}^t$ is the noisy graph at $t$-th diffusion step with $\mathcal{G}^0$ being the original clean graph and $q$ is the noise model. $\mathbf{Q}_{\mathbf{X}}^t$ and $\mathbf{Q}_{\mathcal{E}}^t$ are the transition matrices for $\mathbf{X}$ and $\mathcal{E}$, respectively. 
Following~\cite{vignac2023digress}, with Eq.(\ref{eq:diffusion_one_step_forward_process}), the diffusion from $\mathcal{G}_0$ to $\mathcal{G}_t$ can be written as
\begin{equation}
\label{eq:diffusion_total_process_graph_diffusion_model}
q(\mathcal{G}^t \mid \mathcal{G}^0 )=(\mathbf{X} \overline{{Q}}_\mathbf{X}^t, \mathcal{E} \overline{{Q}}_\mathcal{E}^t),
\end{equation}
where $\overline{{Q}}_\mathbf{X}^t = Q_{\mathbf{X}}^1\ldots Q_{\mathbf{X}}^t$ and $\overline{{Q}}_\mathcal{E}^t = Q_{\mathbf{X}}^1\ldots Q_{\mathcal{E}}^t$. 

In the denoising process, DiGress learns a denoising neural network $R_{\phi}$ parameterized by $\phi$ 
to predict the clean graph $\mathcal{G}^0$ from the noisy graph $\mathcal{G}^t$. 
The denoising process is expressed as:
\begin{equation} \small
\label{eq:denoise_total_process_graph_diffusion_model}
p_\phi(\mathcal{G}^{t-1} \mid \mathcal{G}^t)=\prod_{1 \leq i \leq n} p_\phi(\mathbf{x}_i^{t-1} \mid \mathcal{G}^t) \prod_{1 \leq i, j \leq n} p_\phi(e_{i j}^{t-1} \mid \mathcal{G}^t), 
\end{equation}
where $p_\phi(\mathbf{x}_i^{t-1} \mid \mathcal{G}^t)$ denotes the probability of predicting $\mathbf{x}_i^{t-1}$ from $\mathcal{G}^t$ and $p_\phi(e_{i j}^{t-1} \mid \mathcal{G}^t)$ denotes the probability of predicting $e_{ij}^{t-1}$ from $\mathcal{G}^t$,
where $e_{ij}$ is the edge category between node $i$ and node $j$, 
For simplicity, we define $R_{\phi}$ parameterized by $\phi$ as the function to denoise $\mathcal{G}_{i}$ and then get the next sampled value as 
\begin{equation}
    \mathcal{G}_{i}^{t-1} = R_{\phi}(\mathcal{G}_{i}^{t}, t)
\end{equation}
DiGress is trained to revert the diffusion process to recover the clean graph, which equips the denoising network with the ability of generating realistic graphs from input noises.

\subsection{Threat Model}
\label{sec:threat_model}
\subsubsection{Attacker's Goal} Given a target GNN classifier $f_{{\boldsymbol{\theta}}}$ trained on a private target graph dataset $\mathcal{D}_{t} = \{\mathcal{G}_1,\dots, \mathcal{G}_{|\mathcal{D}_{t}|}\}$, the goal of the adversary in graph stealing attack is to reconstruct graphs in $\mathcal{D}_{t}$ from the target model $f_{{\boldsymbol{\theta}}}$. 

\subsubsection{Attacker's Knowledge and Capability} 
\label{sec:attack_knowledge}
We focus on the graph stealing attacks in the white-box setting. The information of the target model $f_{{\boldsymbol{\theta}}}$ including model architecture and model parameters ${\boldsymbol{\theta}}$ is available to the attacker. This is a reasonable setting in real-world as model providers will often release the trained model to customers for downstream applications. Moreover, we consider a practical setting that the attacker does not have access to any sensitive information of the target dataset $\mathcal{D}_{t}$, including graph topology $\mathbf{A}_i$, node attributes $\mathbf{X}_i$ and graph label $y_i$ of $\mathcal{G}_{i}\in\mathcal{D}_{t}$ as the private data could contain sensitive information or is intellectual property of the data owner. However, we assume that 
% the attackers may have other auxiliary knowledge to facilitate graph stealing attack, such as \minhua{xxx}. Thus, 
an auxiliary dataset $\mathcal{D}_{a}=\{\mathcal{G}_i, y_i\}^{|\mathcal{D}_{a}|}_{i=1}$, which shares a similar low dimensional manifold~\cite{du2020few} as that of $\mathcal{D}_{t}$, is available to the attackers. This assumption is reasonable because in realistic scenarios, e.g., drug design~\cite{vignac2023digress}, the attacker has access to many publicly available molecules and has high motivation to steal private molecules that are easy to synthesize and have high activity on specific targets.

\subsection{Problem Definition}
With the above notations and the description of graph stealing attack in Sec.~\ref{sec:threat_model}, the objective of graph stealing attack is to reconstruct graphs in the target dataset $\mathcal{D}_{t}$ that are used to train the target GNN model without any partial information of $\mathcal{D}_{t}$. This problem can be formally defined as:
\begin{problem}[Graph Stealing Attack]
Given a target GNN classifier: $f_{{\boldsymbol{\theta}}} : \mathcal{G}\xrightarrow{}\mathbf{y} $ trained on a privately owned dataset $\mathcal{D}_{t} = \{(\mathcal{G}_i, y_i)\}^{|\mathcal{D}_{t}|}_{i=1}$, where $\mathbf{y}_i\in \{1,2,\dots,C\}$ is $\mathcal{G}_i$'s label, and an auxiliary dataset $\mathcal{D}_a$ that share the same manifold characteristics as $\mathcal{D}_t$ with $\mathcal{D}_a  \cap \mathcal{D}_t = \emptyset $, we aim to extract a subset of private training data $\mathcal{D}_s \subset \mathcal{D}_t$. The architecture and model parameters of $f_{{\boldsymbol{\theta}}}$ are accessible.  
\end{problem}

\section{Methodology}
\begin{figure}
    \centering
    \includegraphics[width=0.96\linewidth]{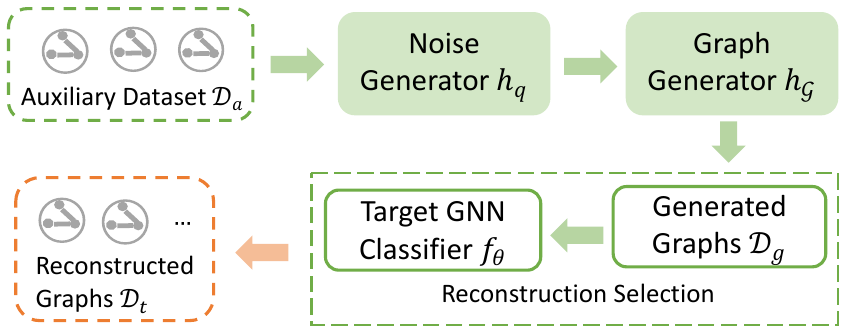}
    \vskip -1.2em
    \caption{An overview of proposed GraphSteal.}
    \vskip -1em
    \label{fig:framework_graphsteal}
\end{figure}

In this section, we present the details of our GraphSteal, which aims to reconstruct the training graphs from the target GNN model without even partial information of the target dataset. There are mainly two challenges to be addressed for achieving better reconstruction performance: \textbf{(i)} how to ensure both the realism and quality of graphs reconstructed from the trained GNN model; and \textbf{(ii)} how to utilize the trained GNN model parameters to extract the training graph information. To address these challenges, we propose a novel framework GraphSteal, which is illustrated in Fig.~\ref{fig:framework_graphsteal}. GraphSteal is composed of a graph generator $h_{\mathcal{G}}$, a noise generator $h_{q}$, a reconstructed graph selector $h_{s}$ and the target GNN model $f_{{\boldsymbol{\theta}}}$. Specifically, a graph diffusion model is adopted as the graph generator $h_{\mathcal{G}}$ to generate realistic and high-quality graphs. The noise generator $h_{q}$ takes graphs from the auxiliary dataset $\mathcal{D}_{a}$ as inputs to learn input noises for the graph generator $h_{\mathcal{G}}$, aiming to ensure $h_{\mathcal{G}}$ can generate graphs $\mathcal{D}_{g}$ mimic to the graphs within the target dataset $\mathcal{D}_t$. $\mathcal{D}_t$ is the training set of $f_{\boldsymbol{\theta}}$. Finally, we perform a reconstruction selection by leveraging parameters $\theta$ of the target classifier $f_{\boldsymbol{\theta}}$ to select the top-$k$ most representative graphs from $\mathcal{D}_{g}$ that closely resemble the target dataset $\mathcal{D}_t$. Next, we give the detailed design of the proposed framework.

\subsection{Reconstruction Generation}
\label{sec:reconstruction_generation}
In this subsection, we give the details of reconstruction generation. We first train a graph diffusion model as the graph generator to guarantee the quality of the reconstructions in Sec.~\ref{sec:building_graph_generator}. We then propose to devise input noises for the graph diffusion model by solving an optimization problem in Sec.~\ref{sec:diffusion_noise_optimization}. Finally, we utilize the optimized input noises to produce the reconstructions in Sec.~\ref{sec:generating_graphs_via_sdedit}.
\subsubsection{Building Graph Generator}
\label{sec:building_graph_generator}
To ensure that we can reconstruct the high-quality and realistic training graphs from trained GNN model, one promising direction is to utilize graph diffusion models that have powerful generation capability to generate graphs. 
Graph diffusion models are usually trained on a specific dataset to learn its distribution. The trained graph diffusion model can then generate diverse realistic samples that are representatives of the learned data distribution. As the auxiliary dataset $\mathcal{D}_a$ share a similar distribution as the target dataset $\mathcal{D}_t$, we first train a graph diffusion model $h_{\mathcal{G}}$ on $\mathcal{D}_a$ to learn the underlying data distribution. 
% as the graph generator $h_{\mathcal{G}}$ to 
After $h_{\mathcal{G}}$ is trained, it can generate high-quality graphs that are likely to follow the distribution of $\mathcal{D}_t$. We adopt DiGress introduced in Sec. ~\ref{sec:graph_diffusion_model} as the graph diffusion model due to its effectiveness.

\subsubsection{Diffusion Noise Optimization}
\label{sec:diffusion_noise_optimization}
Though we can use $h_{\mathcal{G}}$ to generate realistic graphs, there are two issues: (i) If we use random noise as input to $h_{\mathcal{G}}$ to generate graphs, the chance of hitting the training graph in $\mathcal{D}_t$ is very low as the noise input space is very large; and (ii)
Though the auxiliary dataset $\mathcal{D}_a$ shares a similar low dimensional manifold with $\mathcal{D}_t$, they might have a slightly distribution shift from each other. Hence, directly applying $h_{\mathcal{G}}$ trained on $\mathcal{D}_a$  to generate the reconstructed graphs may still not accurately represent the graphs in $\mathcal{D}_t$ due to the distribution shift.  

To address this issue, inspired by~\cite{chen2023importance, meng2022sdedit}, we propose to first find input noises that are likely to results in graphs closely resembling those in $\mathcal{D}_t$, which reformulates the problem of generating training graphs in $\mathcal{D}_t$ as noise optimization problem, i.e., how to design the input noise $\mathcal{G}'$ for the graph diffusion model $h_{\mathcal{G}}$ such that $h_{\mathcal{G}}$ can generate graphs closely resembling those in $\mathcal{D}_t$?

To solve the above problem, we propose a novel diffusion noise generator to learn the input noise to guide the generation of the graph diffusion model to generate graphs better represent those in the target dataset. 
Our intuition is that given a graph $\mathcal{G}_i$ with the ground-truth label $y_i$ , if $\mathcal{G}_i$ has higher probability to be predicted to the target class $y_i$ by the GNN model $f_{\boldsymbol{\theta}}$, $\mathcal{G}_i$ is more likely to be the training graph of $f_{\boldsymbol{\theta}}$. 
Moreover, as the auxiliary dataset is available to the attackers, to improve the efficiency of learning input noise, 
we can then select graphs with higher prediction scores as the input noises to guide the graph diffusion model in generating graphs with similar characteristics to those in target dataset~\cite{meng2022sdedit}.

Therefore, 
we first use a metric to measure the prediction score of the graphs with the target model $f_\theta$ for selecting graphs from $\mathcal{D}_a$ as the input noises for the graph diffusion model $h_{\mathcal{G}}$. 
Formally, given a graph $\mathcal{G}_i\in\mathcal{D}_a$, the metric score is defined as:
\begin{equation}
\label{eq:metric_score_for_selecting_input_graphs}
    s(\mathcal{G}_i,y_i;f_{{\boldsymbol{\theta}}}) = f_{{\boldsymbol{\theta}}}(\mathcal{G}_i)_{y_i}
\end{equation}
where $y_i$ denotes the ground-truth label of $\mathcal{G}_i$. After getting the metric score of each graph in $\mathcal{D}_a$, we select graphs with the top-$m$ highest score in each class. The set of the selected graphs are denoted as $\mathcal{D}_c=\{\mathcal{G}_1,\ldots,\mathcal{G}_{M}\}$, where $M = C\cdot m$.

To further ensure the similarity of selected graphs to the target graphs, thereby improving reconstruction performance,
% that we can obtain the input noise that has similar characteristics with those in the target dataset
we treat the adjacency matrix $\mathbf{A}_i$ and the node attributes $\mathbf{X}_i$ of the graph $\mathcal{G}_i\in\mathcal{D}_c$ as variables to be optimized. Specifically, they are optimized by minimizing the loss between the prediction $f_{\theta}(\mathcal{G}_i)$ and $y_i$ as: 
\begin{equation}
    \label{eq:noise_optimization_objective_function}
    {\sum}_{i=1}^{M}\underset{\mathcal{G}_i\in\mathcal{D}_c}{\min}{\mathcal{L}(f_{{\boldsymbol{\theta}}}(\mathcal{G}_i),y_i)},
\end{equation}
where $\mathcal{L}$ is the training loss (e.g., cross-entropy loss) of $f_{\theta}$.
% $|\mathcal{D}_c|$ is the number of the graphs.  
The optimized graphs are then denoted as $\mathcal{D}'_c = \{\mathcal{G}_1,\ldots,\mathcal{G}_{M}\}$

\subsubsection{Generating Graphs}
\label{sec:generating_graphs_via_sdedit}
After generating optimized graphs based on Sec.~\ref{sec:diffusion_noise_optimization}, one straightforward way of reconstruction is to use these optimized graphs directly as the reconstructed ones. However, such optimized graphs often fall short in terms of realism and validity when applied to real-world scenarios, rendering the reconstruction process ineffective. This shortfall is further evidenced by the experimental results in Sec.~\ref{sec:ablation_studies}, which show the unrealistic nature and invalidity of these optimized graphs. To address this issue, inspired by~\cite{meng2022sdedit,xue2023diffusion}, we propose to apply SDEdit~\cite{meng2022sdedit} to generate graphs to enhance the generation quality and the resemblance of the generated graphs to those within the target dataset. The key idea of SDEdit is to ``hijack'' the generative process of the graph diffusion $h_{\mathcal{G}}$ by adding a suitable amount of noise to smooth out the undesirable details. This noise addition aims to blur out unwanted details while preserving the key structure of the input graphs. The noised graphs are then fed back into $h_{\mathcal{G}}$ to progressively eliminate the noise, yielding denoised outputs that are both realistic and closely resemble the graphs in the target dataset $\mathcal{D}_t$.
Specifically, given an input noise $\mathcal{G}_i\in\mathcal{D}_{c}'$ optimized in Sec.~\ref{sec:diffusion_noise_optimization}, we first diffuse $\mathcal{G}_i$ for $K$ steps to obtain $\mathcal{G}_{i}^{K}$ using diffusing module of $h_{\mathcal{G}}$ as
\begin{equation}
    \mathcal{G}_{i}^{K} \sim q(\mathcal{G}_{i}^{K}|\mathcal{G}_{i})
\end{equation}
where $q(\cdot|\cdot)$ is the noise function in Eq.~(\ref{eq:diffusion_total_process_graph_diffusion_model}). We then run the reverse denoising process 
$R_{\phi}$ in the graph diffusion model $h_{\mathcal{G}}$ according to
Eq.~(\ref{eq:denoise_total_process_graph_diffusion_model}) in Sec.~\ref{sec:graph_diffusion_model} as
\begin{equation}
\label{eq:sdedit_geneartion}
     h_{\mathcal{G}}(\mathcal{G}_i) = R_{\phi}(\dots R_{\phi}(R_{\phi}(\mathcal{G}_{i}^{K}),K-1)\dots,0),
\end{equation}
It bridges the gap between the input data distribution (which is usually close to the target data distribution) and the auxiliary data distribution, ensuring that the generated graphs are realistic and closely similar to the target graphs in $\mathcal{D}_t$. The set of the generated graphs are denoted as $\mathcal{D}_g=\{\mathcal{G}_1,\ldots,\mathcal{G}_M\}$. 

\subsection{Reconstruction Selection}
\label{sec:reconstruction_selection}

With the above process in Sec.~\ref{sec:reconstruction_generation}, we can get a collection of high-quality labeled graphs that follow a similar distribution with the target dataset $\mathcal{D}_t$. However, it is possible that some generated graphs, while resembling the distribution of $\mathcal{D}_t$, do not precisely match the graphs in $\mathcal{D}_t$. Thus, to further refine the performance of our graph stealing attack and enhance the accuracy of our reconstructions, we propose a novel model parameter-guided graph selection method to select the most representative samples among the generated graphs, prioritizing those that most closely approximate the actual graphs in $\mathcal{D}_t$. 
Our major intuition comes from the strong correlation between the model parameters and the training data throughout the training process of neural networks~\cite{Lyu2020Gradient,ji2020directional}. Building upon this insight, we leverage the parameters ${\boldsymbol{\theta}}$ of the target GNN classifier $f_{{\boldsymbol{\theta}}}$ to select the graphs that most closely resemble the target training graphs of $f_{{\boldsymbol{\theta}}}$. Specifically, we first introduce the connection between model parameters and the training data in Sec.~\ref{sec:homogeneous_graph_neural_networks}. Then, we present a novel approach based on the above connection to select samples from $\mathcal{D}_g$ that most closely resemble to graphs in the target dataset $\mathcal{D}_t$ in Sec.~\ref{sec:optimizing_selection_mask}.

\subsubsection{Connections between Model Parameters and Training Data}
\label{sec:homogeneous_graph_neural_networks}
% We begin by providing a formal definition of homogeneous neural networks:
We begin by formally defining homogeneous neural networks:
\begin{definition}[Homogeneous Neural Networks~\cite{Lyu2020Gradient}]
    \label{def:homogeneous_neural_network}
    Let $f$ be a neural network with model parameters as ${\boldsymbol{\theta}}$. Then $f$ is a \textit{homogeneous} neural network if there is a number $L>0$ such that the model output $f(\mathcal{G};{\boldsymbol{\theta}})$ satisfies the following equation:
    \begin{equation}
    \label{eq:homogeneous_neural_networks}
    f(\mathcal{G};\sigma{\boldsymbol{\theta}})=\sigma^{L}f(\mathcal{G};{\boldsymbol{\theta}}),~~\forall{\sigma>0}
    \end{equation}
\end{definition}

Note that essentially any message-passing based GNNs (e.g. GCN~\cite{kipf2016semi} and SGC~\cite{wu2019simplifying}) with ReLU activations is homogeneous w.r.t the parameters ${\boldsymbol{\theta}}$ if it does not have any skip-connections (e.g., GraphSage~\cite{hamilton2017inductive}) or bias terms, except possibly for the first layer. More details of the proof and discussion is in Appendix~\ref{sec:homogeneity_GNN_appendix}.

We then present Lemma~\ref{lemma:gradient_descent} below to show the connection between model parameters and the training data of homogeneous neural networks. Essentially, the lemma shows that the process of training GNNs with gradient descent or gradient flow can be formulated as a constrained optimization problem related to margin maximization, which is formally given as:
\begin{lemma}[\cite{Lyu2020Gradient}]
    \label{lemma:gradient_descent}
    Let $f_{{\boldsymbol{\theta}}}$ be a homogeneous ReLU neural network with model parameters ${\boldsymbol{\theta}}$. Let $\mathcal{D}_t=\{(\mathcal{G}_{i},y_{i})\}^{|\mathcal{D}_t|}_{i=1}$ be the classification training dataset of $f_{{\boldsymbol{\theta}}}$, where $y_i\in\{1,\ldots,C\}$. Assume $f_{{\boldsymbol{\theta}}}$ is trained by minimizing the cross entropy loss 
    % $\mathcal{L}_{ce} = \sum_{i=1}^{|\mathcal{D}_t|}y_i\cdot \log{}$
    $\mathcal{L}_{ce} = \sum_{i=1}^{|\mathcal{D}_t|}\log({1+\sum_{j\neq y_i}e^{-q_{ij}(\boldsymbol{\theta})}})$ over $\mathcal{D}_t$ using gradient flow, where $q_{ij}(\boldsymbol{\theta}) = f_{\boldsymbol{\theta}}(\mathcal{G}_i)_{y_i} - f_{\boldsymbol{\theta}}(\mathcal{G}_i)_{j}$ and $f_{\boldsymbol{\theta}}(\mathcal{G}_i)_{y_i}$ denotes the $y_i$-th entry of the logit score vector $f_{\boldsymbol{\theta}}(\mathcal{G}_i)$ before applying the softmax normalization. Then, gradient flow converges in direction towards a first-order stationary point of the following max-margin optimization problem:
    \begin{equation}
        \label{eq:max_margin_optimization}
        \underset{{{\boldsymbol{\theta}}}}{\min}\frac{1}{2}||{{\boldsymbol{\theta}}}||^2_2, \quad \text{s.t.} \quad 
        q_{ij}(\boldsymbol{\theta}) \geq 1,~~\forall{i\in[|\mathcal{D}_t|]}, {j\in[C]\backslash \{y_i\}}. 
    \end{equation}
\end{lemma}
Lemma~\ref{lemma:gradient_descent} guarantees directional convergence to a \textit{Karush-Kuhn-Tucker} point (or \textit{KKT} point). More details of KKT points can be found in Appendix~\ref{appendix:kkt_points}. 

As described in Sec.~\ref{sec:homogeneous_graph_neural_networks}, we can extend Lemma~\ref{lemma:gradient_descent} to homogeneous GNNs. Formally, given a target GNN classifier $f_{{\boldsymbol{\theta}}}$ that satisfy Eq.~(\ref{eq:homogeneous_neural_networks}) in Def.~\ref{def:homogeneous_neural_network}, the formal theorem is presented as follows:
\begin{theorem}
\label{thm:relationship_parameter_training_data}
Let $f_{{\boldsymbol{\theta}}}$  be the target GNN classifier that preserves the homogeneity in Def.~\ref{def:homogeneous_neural_network}. $f_{{\boldsymbol{\theta}}}$ is trained on the labeled set $\mathcal{D}_t=\{(\mathcal{G}_{i},y_{i})\}^{|\mathcal{D}_t|}_{i=1}$, where $y_i\in\{1,\ldots,C\}$. 
After convergence of training $f_{{\boldsymbol{\theta}}}$ using gradient descent, the converged model parameter $\tilde{\boldsymbol{\theta}}$ satisfies the following equation:
\begin{equation}
\label{eq:kkt_stationary}
        \tilde{{\boldsymbol{\theta}}} = \sum_{i=1}^{|\mathcal{D}_t|}\lambda_{i}(\nabla_{{\boldsymbol{\theta}}}f_{\tilde{{\boldsymbol{\theta}}}}(\mathcal{G}_i)_{y_i} - \nabla_{{\boldsymbol{\theta}}}\max_{j\neq y_i}\{f_{\tilde{{\boldsymbol{\theta}}}}(\mathcal{G}_i)_{j}\}),
\end{equation}
where
\begin{align}
    & \beta_{i}(\tilde{\boldsymbol{\theta}}) = f_{\tilde{\boldsymbol{\theta}}}(\mathcal{G}_i)_{y_i} - \max_{j\neq y_i}\{f_{\tilde{{\boldsymbol{\theta}}}}(\mathcal{G}_i)_{j}\}\geq 1, \forall {i\in |\mathcal{D}_t|}, \\
         &\lambda_1,\ldots,\lambda_{|\mathcal{D}_t|} \geq 0, \\
       & \lambda_i=0 \quad \text{if}~~\beta_{i}(\tilde{\boldsymbol{\theta}})\neq{1}, \forall {i\in |\mathcal{D}_t|}. \label{eq:complementary_slackness}
\end{align}
\end{theorem}

The proof is in Appendix~\ref{appendix:proof_of_thm_relationship_parameter_training_data}. 
Eq.~(\ref{eq:kkt_stationary}) implies that the parameters $\tilde{{\boldsymbol{\theta}}}$ are a linear combination of the derivatives of the target GNN $f_{{\boldsymbol{\theta}}}$ at the training graphs $\mathcal{G}_i\in\mathcal{D}_t$. Especially, according to Eq.~(\ref{eq:complementary_slackness}), it is possible that $\lambda_i = 0$ for some training graph $\mathcal{G}_i\in\mathcal{D}_t$, which implies that such $\mathcal{G}_i$ has no impact on the final model parameters of $f_{\boldsymbol{\theta}}$. Thus, to recover training dataset $\mathcal{D}_t$ from $f_{\boldsymbol{\theta}}$, it is natural  to identify  graphs $\{\mathcal{G}_1, \ldots, \mathcal{G}_k\}$ 
% from $\mathcal{G}_i\in\mathcal{D}_{g}$ 
from $\mathcal{D}_{g}$ with $\lambda_i > 0$ to satisfy Eq.~(\ref{eq:kkt_stationary}), which means they contribute to $\boldsymbol{\theta}$.

\subsubsection{Optimizing Selection Mask}
\label{sec:optimizing_selection_mask}
Based on Theorem~\ref{thm:relationship_parameter_training_data}, the selection of the reconstructed graphs can be reformulated as finding optimal graph selection masks $\Lambda=\{\lambda_1,\ldots,\lambda_{M}\}$ for the set of generated graphs $\mathcal{D}_{g} = \{\mathcal{G}_{1},\ldots,\mathcal{G}_{M}\}$ in Sec.~\ref{sec:generating_graphs_via_sdedit}. Specifically, our objective is to optimize $\Lambda$ to select graphs from $\mathcal{D}_{g}$ to closely approximate the model parameters $\boldsymbol{\theta}$ for $f_{\boldsymbol{\theta}}$. The objective function for selection can be written as:
\begin{equation} \small
\begin{aligned}
\label{eq:selection_mask_objective}
    & \min_{\Lambda} \mathcal{L}_{s}(\mathcal{D}_{g}, \Lambda, \boldsymbol{\theta}) = \left\| \boldsymbol{\theta} - \sum_{i=1}^{M} \lambda_{i}\left(\nabla_{{\boldsymbol{\theta}}}f_{{{\boldsymbol{\theta}}}}(\mathcal{G}_i)_{y_i} - \nabla_{{\boldsymbol{\theta}}}\max_{j\neq y_i}\{f_{{\boldsymbol{\theta}}}(\mathcal{G}_i)_j\}\right) \right\|_2^2, \\
    & \text{s.t.} \quad \lambda_{i}\geq 0, \quad \forall i \in \{1, \dots, M\}
\end{aligned}
\end{equation}
where $\boldsymbol{\theta}$ and $\mathcal{G}_i\in\mathcal{D}_{g}$ are known and $\{\lambda_1,\dots,\lambda_{M}\}$ is the set of graph selection mask to be optimized. 
$\lambda_i$ can be view as the metric score of selecting $\mathcal{G}_i$ as the final reconstructed graphs. Since this is a constrained optimization problem, to fulfill the constraint in Eq.~(\ref{eq:selection_mask_objective}), we introduce a constraint loss $\mathcal{L}_\lambda(\Lambda)$, which is defined as:
\begin{equation}
\label{eq:contraint_loss}
\mathcal{L}_\lambda(\Lambda) = {\sum}_{i=1}^{M} \max\{-\lambda_i,0\}.
\end{equation}
Therefore, the final objective function for selection is:
\begin{equation}
\label{eq:final_objective_function}
    \min_{\Lambda} \mathcal{L}_{s}(\mathcal{D}_{g}, \Lambda, \boldsymbol{\theta}) + \alpha\mathcal{L}_\lambda(\Lambda),
\end{equation}
where $\alpha$ is the hyperparameter to balance the contribution of $\mathcal{L}_\lambda(\Lambda)$. After $\mathcal{L}_\lambda(\Lambda)$ is learned, we select graphs with the top-$k$ highest scores as the final reconstructed graphs, which are denoted as $\mathcal{D}_r=\{\mathcal{G}_1,\ldots,\mathcal{G}_{k}\}$.

\subsection{Reconstruction Algorithm}
The reconstruction algorithm of GraphSteal is in Alg.~\ref{alg:reconstructioin_alg}. Specifically, we train a graph diffusion model $h_{\mathcal{G}}$ as the graph generator (line 1), and select graphs from the auxiliary dataset $\mathcal{D}_a$ as the input graphs $\mathcal{D}_c$ (line 2). $\mathcal{D}_c$ is then optimized based on Eq.~(\ref{eq:noise_optimization_objective_function}) (line 3) to use to generate $\mathcal{D}_g$ based on Eq.~(\ref{eq:sdedit_geneartion}) (line 4). From line 5 to line 8, we learn the graph selection masks $\Lambda$ based on Eq.~(\ref{eq:selection_mask_objective}). Finally, we select graphs with the top-$k$ highest value in $\Lambda$ as $\mathcal{D}_r$ (line 9).

\begin{algorithm}[t] 
\caption{Reconstruction Algorithm of GraphSteal} 
\begin{algorithmic}[1]
\REQUIRE Auxiliary dataset $\mathcal{D}_a$, trained target GNN classifier $f_{\boldsymbol{\theta}}$ with parameters $\boldsymbol{\theta}$, hyperparameter $\alpha$ and the selection number $k$
\ENSURE Reconstructed graphs set $\mathcal{D}_{r}$

\STATE Train a graph generator $h_{\mathcal{G}}$ on $\mathcal{D}_a$
\STATE Select input graphs from $\mathcal{D}_a$ based on Eq.~(\ref{eq:metric_score_for_selecting_input_graphs}) and obtain $\mathcal{D}_c$ 
\STATE Optimize $\mathcal{D}_c$ to get $\mathcal{D}'_c$ based on Eq.~(\ref{eq:noise_optimization_objective_function}) 
\STATE Generate graphs $\mathcal{D}_{g}$ based on Eq.~(\ref{eq:sdedit_geneartion})
\WHILE{not converged}
    \STATE Calculate $\mathcal{L}_{s}(\mathcal{D}_{g}, \Lambda, \boldsymbol{\theta})$ based on Eq.~(\ref{eq:selection_mask_objective})
    \STATE Calculate $\mathcal{L}_{\lambda}(\Lambda)$ based on Eq.~(\ref{eq:contraint_loss}).
    \STATE Update the selection masks $\Lambda$ with gradient decent on $\nabla_{\Lambda}(\mathcal{L}_{s}+\alpha\mathcal{L}_{\lambda})$ based on Eq.~(\ref{eq:final_objective_function}).
\ENDWHILE
\STATE Select graphs with the top-$k$ highest value in $\Lambda$ as $\mathcal{D}_r$
\label{alg:reconstructioin_alg}
\end{algorithmic}
\end{algorithm}

% \newpage
\section{Experiments}
In this section, we conduct experiments on various real-world datasets to answer the following research questions: (i) \textbf{Q1}: Can GraphSteal effectively reconstruct private training graphs of GNNs by leveraging the parameters of GNNs? (ii) \textbf{Q2}: How do the number of final selected graphs affect the performance of reconstruction? (iii) \textbf{Q3}: How does each component of GraphSteal contribute to the effectiveness in stealing training graphs?

\subsection{Experimental Setup}
\subsubsection{Datasets}
We conduct experiments on 3 public real-world datasets, i.e., FreeSolv, ESOL and QM9~\cite{wu2018moleculenet}. FreeSolv and ESOL are small-scale molecular datasets. QM9 is a large-scale-molecular dataset. For each experiment, 
$20\%$ of randomly selected graphs from the original dataset serves as the target dataset for training the target GNN classifier. Additionally, another $10\%$ of the graphs are set as the validation set, while the remaining $70\%$ are assigned as the test set. Note that we set the test set as the auxiliary datase for attackers to train the graph generator and conduct reconstruction.
The statistics of the datasets are summarized in Tab.~\ref{tab:dataset_statistics}. 
More details of the dataset settings are shown in \textit{Appendix~\ref{appendix:datasets}}.

\subsubsection{Baselines}
To the best of our knowledge, GraphSteal is the first graph stealing attack to extract training graphs without access to the private graph-level dataset. To demonstrate the effectiveness of GraphSteal, we first propose three variants: 
\begin{itemize}[leftmargin=*]
    \item \textbf{BL-Rand}: It randomly generates graphs using Erdos-Renyi (ER) models, where node features are randomly sampled from the graphs in the auxiliary dataset.
    \item \textbf{BL-Conf}: This method directly selects the top-$k$ most confidence graphs from the auxiliary dataset as the reconstructed graphs.
    \item  \textbf{BL-Diff}: It applies the graph diffusion model trained on the auxiliary datasets to generate graphs without any modification and selection.
\end{itemize}
Moreover, we compare GraphSteal with a state-of-the-art model inversion attack method for graph neural networks:
\begin{itemize}[leftmargin=*]
    \item  \textbf{GraphMI-G}: This method is extended from \textit{GraphMI}~\cite{zhang2021graphmi}, which is designed for node classification. Specifically, GraphMI aims to reconstruct the adjacency matrix of the target graphs based on the node features and labels of the target graphs. To adapt it to our setting, where there is no partial information about the target dataset, we randomly select graphs from the auxiliary dataset as input graphs. We then apply GraphMI to modify the adjacency matrices of these graphs. The modified graphs are finally regarded as the reconstructed graphs.
\end{itemize}
Additionally, we also consider a state-of-the-art model-level explanation method as the baseline:
\begin{itemize}[leftmargin=*]
    \item \textbf{XGNN}~\cite{yuan2020xgnn}: It generates representative graphs for each class that the target classifier is most confident with as model-level explanations. To adapt it to our setting, we treat these generated representative graphs as reconstructed training graphs.
\end{itemize}
\begin{table}[t]
    \centering
    \caption{Statistics of Datasets}
    \small
    \vskip -1em 
    \begin{tabularx}{0.95\linewidth}{p{0.12\linewidth}p{0.12\linewidth}XXXX}
    \toprule 
    Datasets & \#Graphs & \#Avg. Nodes & \#Avg. Edges & \#Avg. Feature & \#Classes \\%& \#Attack Budget\\
    \midrule
    FreeSolv & 642 & 8.7 & 16.8 & 9 & 2 \\
    ESOL & 1,128 & 13.3 & 27.4 & 9 & 2 \\ %& 10\\
    QM9 & 130,831 & 8.8 & 18.8 & 11 & 3 \\ %& 40\\

    \bottomrule
    \end{tabularx}
    \vskip -1em
    \label{tab:dataset_statistics}
\end{table}

\begin{table*}[t]
    \centering
    % \small
    %\vskip -1em
    \caption{Comparison with baselines in stealing training graphs on various graph datasets against GCN.}
    \vskip -0.8em
    \resizebox{0.9\textwidth}{!}
    {\begin{tabularx}{0.98\linewidth}{p{0.08\linewidth}p{0.13\linewidth}CCCCCc}
    \toprule
    {Dataset} & 
    {Metrics} & {BL-Rand}& {BL-Conf}& {BL-Diff}& {XGNN}& {GraphMI-G}& {GraphSteal}\\
     \midrule
\multirow{4}{*}{FreeSolv}
    & Validity (\%) $\uparrow$ & 20.3$\pm$5.4& \textbf{100$\pm$0} & 99.3$\pm$0.8 & 17.0$\pm$2.5 & 33.0$\pm$1.2 &{98.8$\pm$0.8} \\

& Uniqueness (\%)  $\uparrow$ & {97.4$\pm$3.6} & 95.6$\pm$0.3 & 77.0$\pm$3.2& \textbf{100$\pm$0} & 66.7$\pm$0.5 & 81.8$\pm$1.8\\

& Recon. Rate (\%)  $\uparrow$ & 4.0$\pm$3.4 & 2.2$\pm$0.8 & 39.4$\pm$4.2 & 1.9$\pm$1.8 & 4.5$\pm$0.9 &\textbf{50.4$\pm$2.3} \\
& FCD $\downarrow$ &  16.9$\pm$1.2 & 9.5$\pm$0.9 & 6.6$\pm$0.3 &23.4$\pm$1.1 & 16.6$\pm$0.5 & \textbf{5.7$\pm$0.5}\\
\midrule
\multirow{4}{*}{ESOL}
& Validity (\%)$\uparrow$ & 15.0$\pm$2.8 & \textbf{100$\pm$0} & 98.7$\pm$1.2 & 14.7$\pm$1.7 & 14.3$\pm$0.5 & 96.7$\pm$2.2 \\

& Uniqueness (\%)$\uparrow$ & \textbf{100$\pm$0} & 95.0$\pm$0.4 & 94.5$\pm$0.9 & 100$\pm$0 & 86.0$\pm$0.5 & 95.6$\pm$0.8\\

& Recon. Rate (\%) $\uparrow$ & 0$\pm$0 & 0$\pm$0 & 4.8$\pm$0.8 & 0$\pm$0 & 0$\pm$0 & \textbf{8.6$\pm$1.4} \\
& FCD $\downarrow$ &  27.0$\pm$2.7 & 17.6$\pm$0.1 & {4.9$\pm$0.5} & 29.9$\pm$0.5 & 27.8$\pm$7.9 & \textbf{4.5$\pm$0.3}\\
\midrule
\multirow{4}{*}{QM9}
& Validity (\%) $\uparrow$ & 3.3$\pm$0.5 & \textbf{100$\pm$0} & 98.4$\pm$1.7& 57.0$\pm$1.6 & 4.7$\pm$0.5 & {98.8$\pm$0.8}\\

& Uniqueness (\%) $\uparrow$ & \textbf{100$\pm$0} & 82.6$\pm$3.1& 98.2$\pm$2.7& 98.2$\pm$1.5 & \textbf{100$\pm$0} & \textbf{100$\pm$0}\\
    
& Recon. Rate (\%) $\uparrow$ & 0$\pm$0 & 1.9$\pm$0.9 & 19.4$\pm$2.8 & 0$\pm$0 & 0$\pm$0 & \textbf{29.2$\pm$3.1}\\
& FCD $\downarrow$ & 21.6$\pm$2.3& 9.6$\pm$0.8 & 2.8$\pm$1.8 & 12.2$\pm$0.5 & 19.9$\pm$1.1 & \textbf{2.0$\pm$0.2}\\
    \bottomrule 
        \end{tabularx}}
    % \vskip -1.em
    \label{tab:comparisons_with_baselines}
\end{table*}

\subsubsection{Implementation Details}
\label{sec:implementation_details}
In this paper, we conduct experiments on the inductive supervised graph classification task, 
\textit{where the adversary aims to extract training graphs from the privately owned dataset}. To demonstrate the transferability of GraphSteal, we target GNNs with various architectures, i.e., 2-layer GCN~\cite{kipf2016semi}, 2-layer GIN~\cite{xu2018powerful} and 9-layer Graph Transformer (GTN)~\cite{dwivedi2021generalization}. DiGress~\cite{vignac2023digress} is set as the basic graph diffusion model to generate graphs.
All hyperparameters of the compared methods are tuned for fair comparisons. Each experiment is conducted $5$ times on an A6000 GPU with 48G memory and the average results are reported. 
\subsubsection{Evaluation Metrics} 
\label{sec:evaluation_protocol}
\textit{First}, we adopt the following metrics  to evaluate the \textit{realism} and \textit{quality} of the reconstructed graphs:
\begin{itemize}[leftmargin=*]
    \item \textbf{Validity} calculates the fraction of the reconstructed molecular graphs adhering to fundamental chemical rules and principles as $\text{Validity} = V(\mathcal{D}_r)/|\mathcal{D}_r|$, where $V(\mathcal{D}_r)$ is the number of graphs in $\mathcal{D}_r$ adhering to fundamental chemical rules and principles measured by RDKit~\footnote{https://www.rdkit.org/}. A larger validity score implies the better quality and realism of the reconstructed graphs.
    \item \textbf{Uniqueness} measures the proportion of graphs that exhibit uniqueness within the entire set of reconstructed graphs, which is calculated by $\text{Uniqueness} = U(\mathcal{D}_r) / |\mathcal{D}_r|$, where $U(\mathcal{D}_r)$ is the number of unique graphs within $\mathcal{D}_r$. A larger uniqueness score implies a better diversity in reconstruction.
\end{itemize}
\textit{Second}, we apply the following metrics to evaluate the \textit{fidelity} of the reconstructed graphs in mirroring the graphs in the target dataset:
\begin{itemize}[leftmargin=*]
    \item \textbf{Reconstruction Rate} represents the fraction of reconstructed graphs matching those in the target dataset as Reconstruction rate $= |\mathcal{D}_r \cap \mathcal{D}_t|/|\mathcal{D}_r|$, where $|\mathcal{D}_r \cap \mathcal{D}_t|$ denotes the number of reconstructed graphs exactly matched with those in $\mathcal{D}_t$. A larger reconstruction rate indicates a better reconstruction method. More details of the exact graph matching are in Appendix~\ref{appendix:additional_details_graph_matching}.
    \item \textbf{Fréchet ChemNet Distance (FCD)}~\cite{preuer2018fcd} quantifies the distance between the distributions of the representations of the reconstructed dataset and the target dataset. The representations are learned from the pre-trained ChemNet~\cite{goli2020chemnet} neural networks. More details of the computation process of FCD is in Appendix~\ref{appendix:FCD}. A lower FCD indicates a better reconstruction method. 
\end{itemize}

\subsection{Reconstruction Results}
To answer $\textbf{Q1}$, we compare GraphSteal with baselines in reconstructing private graphs from the target training dataset on three molecular graph datasets. We also evaluate the performance of GraphSteal against various GNN models to validate its flexibility. Moreover, the impacts of the training/auxiliary set distribution shift and split ratio are further investigated in  \textit{Appendix~\ref{sec:impact_ta}}. 
\subsubsection{Comparisons with Baselines}
\label{sec:comparisions_with_baselines}
We focus on attacking the GNNs for the graph classification task. The target GNN is set as GCN with a sum pooling layer. The final selection number is set as 100. The average reconstruction results on FreeSolv, ESOL, and QM9 are reported in Tab.~\ref{tab:comparisons_with_baselines}. 
Note that a lower FCD indicates a better reconstruction performance. From Tab.~\ref{tab:comparisons_with_baselines}, we can observe that:
\begin{itemize}[leftmargin=*]
    \item Existing baselines give poor performance in validity, reconstruction rate and FCD among all datasets. It implies the necessity of developing graph stealing attacks for reconstructing realistic and high-quality graphs from the privately owned training dataset.
    \item GraphSteal gives superior validity and uniqueness than baselines. 
    It indicates the effectiveness of GraphSteal in reconstructing realistic graphs that are valid in real-world scenarios even on the large-scale dataset QM9. 
    \item GraphSteal shows a much better reconstruction rate and FCD than baselines. This demonstrates GraphSteal can obtain high-quality reconstructions by generating and selecting graphs that closely match those in the target dataset.
\end{itemize}

\subsubsection{Flexibility to Model Architecture}
\label{sec:flexibility_to_model_arch}
\begin{figure}[t]
    \small
    \centering
    \begin{subfigure}{0.49\linewidth}
        \includegraphics[width=0.98\linewidth]{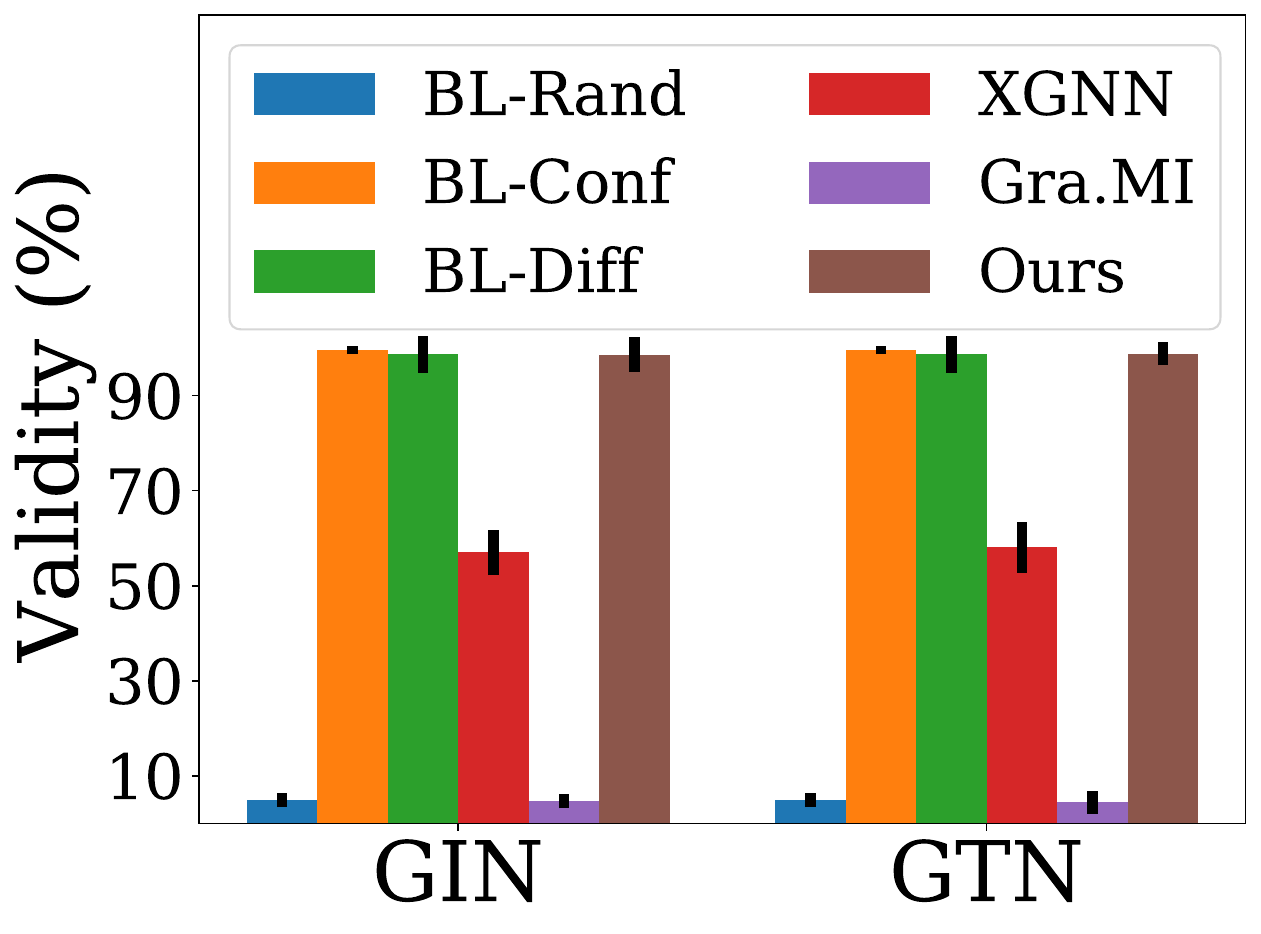}
        \vskip -0.5em
        \caption{Validity}
    \end{subfigure}
    \begin{subfigure}{0.49\linewidth}
        \includegraphics[width=0.98\linewidth]{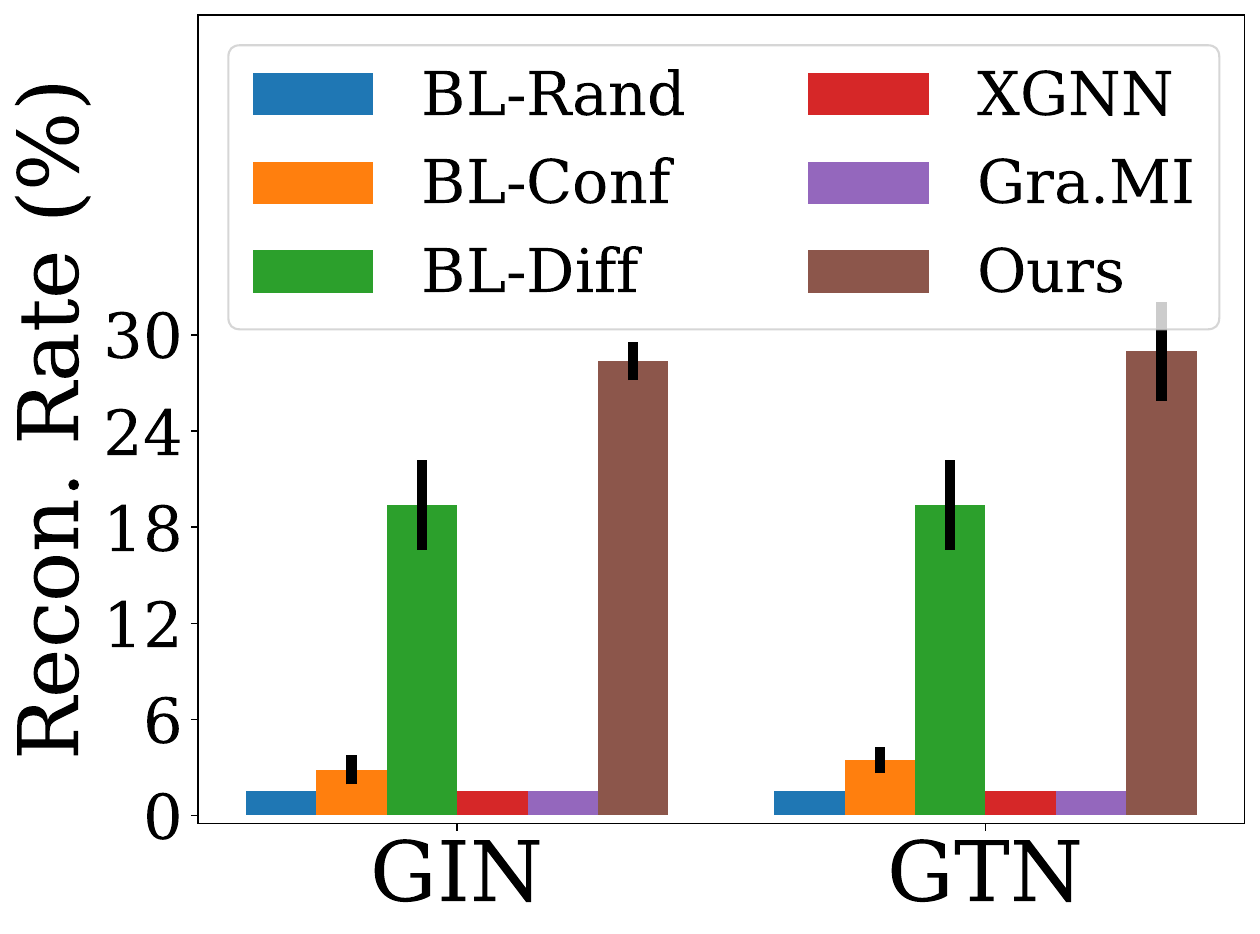}
        \vskip -0.5em
        \caption{Recon. Rate}
    \end{subfigure}
    \vskip -1.5em
    \caption{Reconstruction results on QM9 for various GNNs.}
    \vskip -1em
    \label{fig:flexibility_QM9}
\end{figure}

To show the flexibility of GraphSteal to different GNN models, we consider two more GNNs, i.e., GIN~\cite{xu2018powerful} and GTN~\cite{dwivedi2021generalization}, as the target models to conduct graph stealing attack. The number of selected reconstructed graphs is set as $100$. All other settings are the same as that in Sec.~\ref{sec:implementation_details}. We compare GraphSteal with baselines on QM9. The average results of validity, and reconstruction rate are reported in Fig.~\ref{fig:flexibility_QM9}. 
Similar trends are also observed on other datasets.
% , which can be found in Appendix~\ref{}.
% More results on other datasets can be found in Appendix~\ref{}. 
From the figure, we observe that GraphSteal consistently achieves better reconstruction rate, while maintaining high validity, compared to baselines for both GIN and GTN models. It demonstrates the effectiveness of GraphSteal in reconstructing realistic and high-quality graphs for various GNN models, which shows the flexibility of GraphSteal to steal graphs from various GNN classifiers.

\subsection{Impact of the Number of Selected Graphs}
\label{sec:impact_of_the_number_of_selected_graphs}
\begin{figure}[t]
    \small
    \centering
    \begin{subfigure}{0.46\linewidth}
        \includegraphics[width=0.98\linewidth]{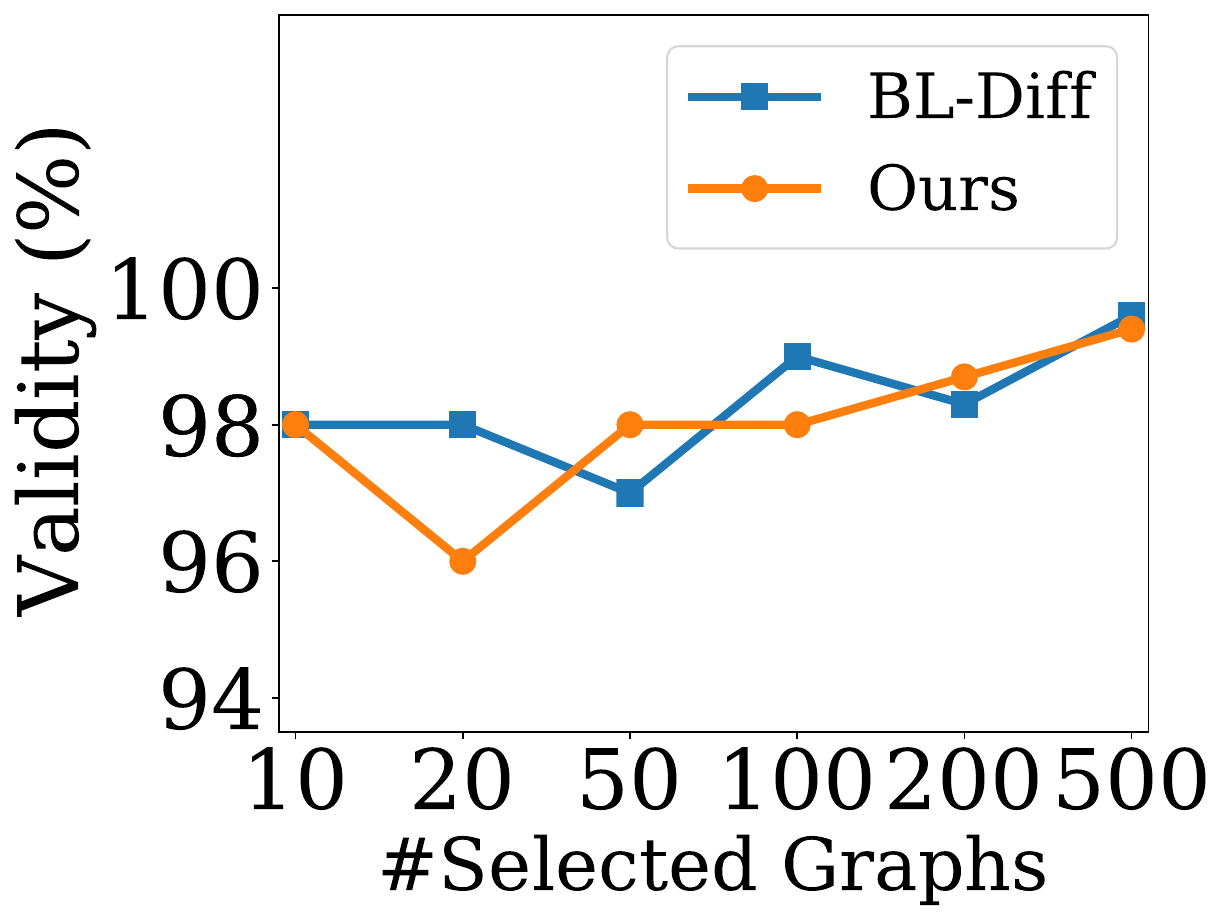}
        \vskip -0.5em
        \caption{Validity}
    \end{subfigure}
    \begin{subfigure}{0.46\linewidth}
        \includegraphics[width=0.98\linewidth]{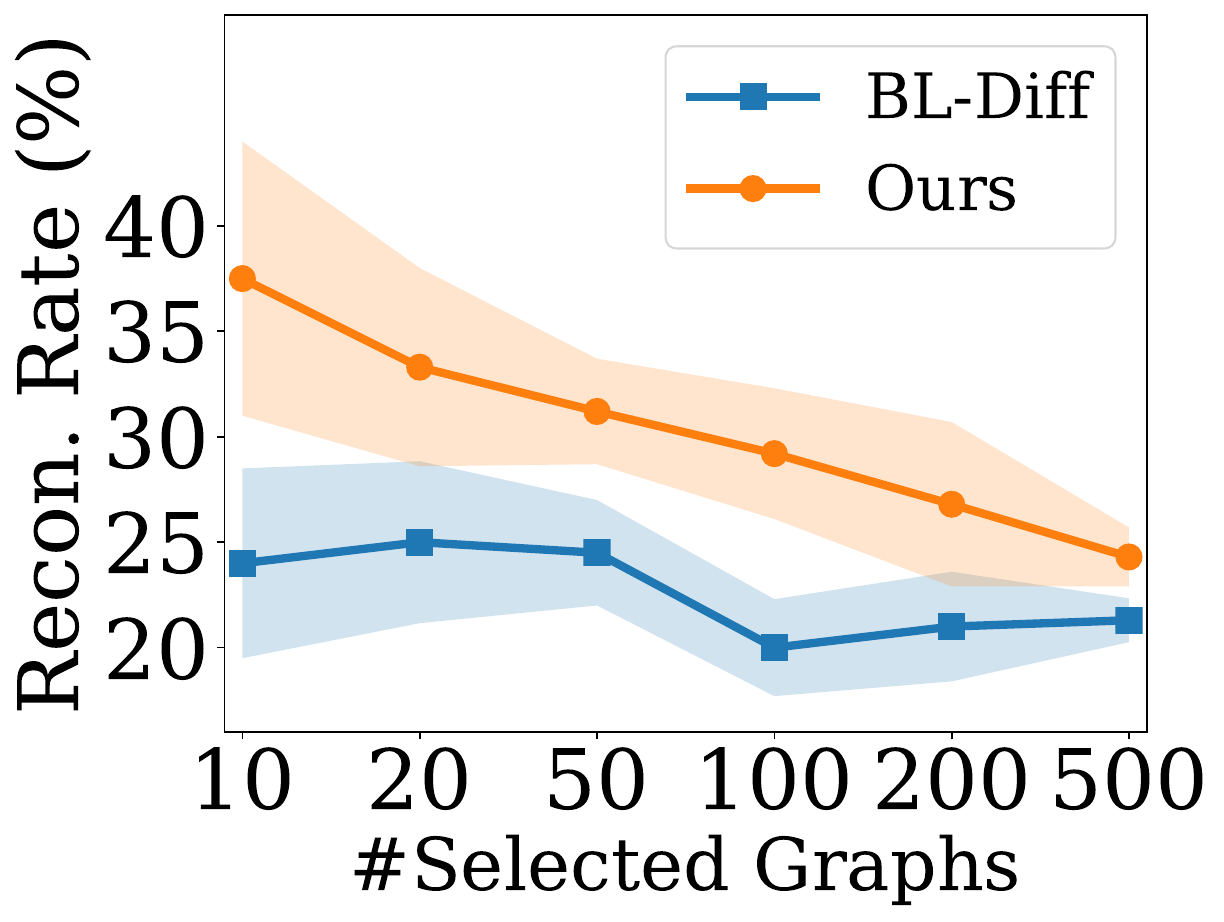}
        \vskip -0.5em
        \caption{Recon. Rate}
    \end{subfigure}
    \vskip -1.5em
    \caption{Impact of the numbers of selected graphs on QM9}
    \vskip -1em
    \label{fig:impact_numbers_of_selected_graphs}
\end{figure}

To answer $\textbf{Q2}$, we conduct experiments to explore the attack performance of GraphSteal given different budgets in the numbers of selected graphs after reconstruction. Specifically, we vary the number of selected graphs as $\{10, 20, 50, 100, 200, 500\}$. GCN is set as the target GNN model. The other settings are the same as Sec.~\ref{sec:implementation_details}. Fig.~\ref{fig:impact_numbers_of_selected_graphs} reports the validity and reconstruction rate on QM9 datasets. More results on other datasets can be found in \textit{Appendix~\ref{appendix:additional_impact_selected_graphs}}.
From the figure, we observe that:
\begin{itemize}[leftmargin=*]
    \item As the number of selected graphs increases, the reconstruction rate of our method would slightly decrease. However, it is worth noting that the absolute number of the matched reconstructed graphs (\# Selected Graphs $\times$ Recon. Rate) increases, which satisfies our expectation. This is because according to Theorem~\ref{thm:relationship_parameter_training_data}, the trained GNN classifier $f_{\boldsymbol{\theta}}$ is regarded as a weighted combination of a subset of the derivative of training graphs. Thus,  there is a theoretical upper bound to the number of graphs that can be reconstructed based on $\boldsymbol{\theta}$. Moreover, our proposed reconstruction selection method can effectively identify the reconstructed graphs that are more likely to be training graphs, leading to a high reconstruction rate when the selected graph number is small. As the number of selected graphs increases, the number of matched reconstructed graphs will gradually reach its upper bound, resulting in a diminished reconstruction rate. More details of the absolute number of matched reconstructed graphs are in \textit{Appendix~\ref{appendix:additional_impact_selected_graphs}}.
    \item Our method consistently outperforms BL-diff in terms of reconstruction rate as the number of selected graphs increases, while still maintaining high validity, which demonstrates the effectiveness of our method in reconstructing realistic and high-quality graphs in real-world scenarios.
\end{itemize}

\subsection{Ablation Study}
\label{sec:ablation_studies}
To answer $\textbf{Q3}$, we conduct ablation studies to understand the effects of the reconstruction generation and reconstruction selection method. (i) To demonstrate the effectiveness of the diffusion noise optimization in Sec.\ref{sec:diffusion_noise_optimization}, we implement a variant GraphSteal/O that replaces the optimized graphs with the Gaussian noise as the input to the graph diffusion model. (ii) GraphSteal deploys a graph diffusion model to improve the realism and quality of reconstruction. To prove its effectiveness, we directly regard the optimized graphs in Sec.\ref{sec:diffusion_noise_optimization} as the reconstructed graphs and train a variant named GraphSteal/D. (iii) A variant named GraphSteal/S is trained by replacing the reconstruction selection component with the random selection. (iv) We also propose a variant named GraphSteal/G by directly conducting selection from the auxiliary dataset based on Eq.~(\ref{eq:selection_mask_objective}) without generating graphs to show the effectiveness of the overall reconstruction generation. BL-Diff is also adopted as a reference. GCN is set as the target GNN model. The number of selected graphs is set as $100$. 

\begin{figure}[t]
    \small
    \centering
    \begin{subfigure}{0.42\linewidth}
        \includegraphics[width=1\linewidth]{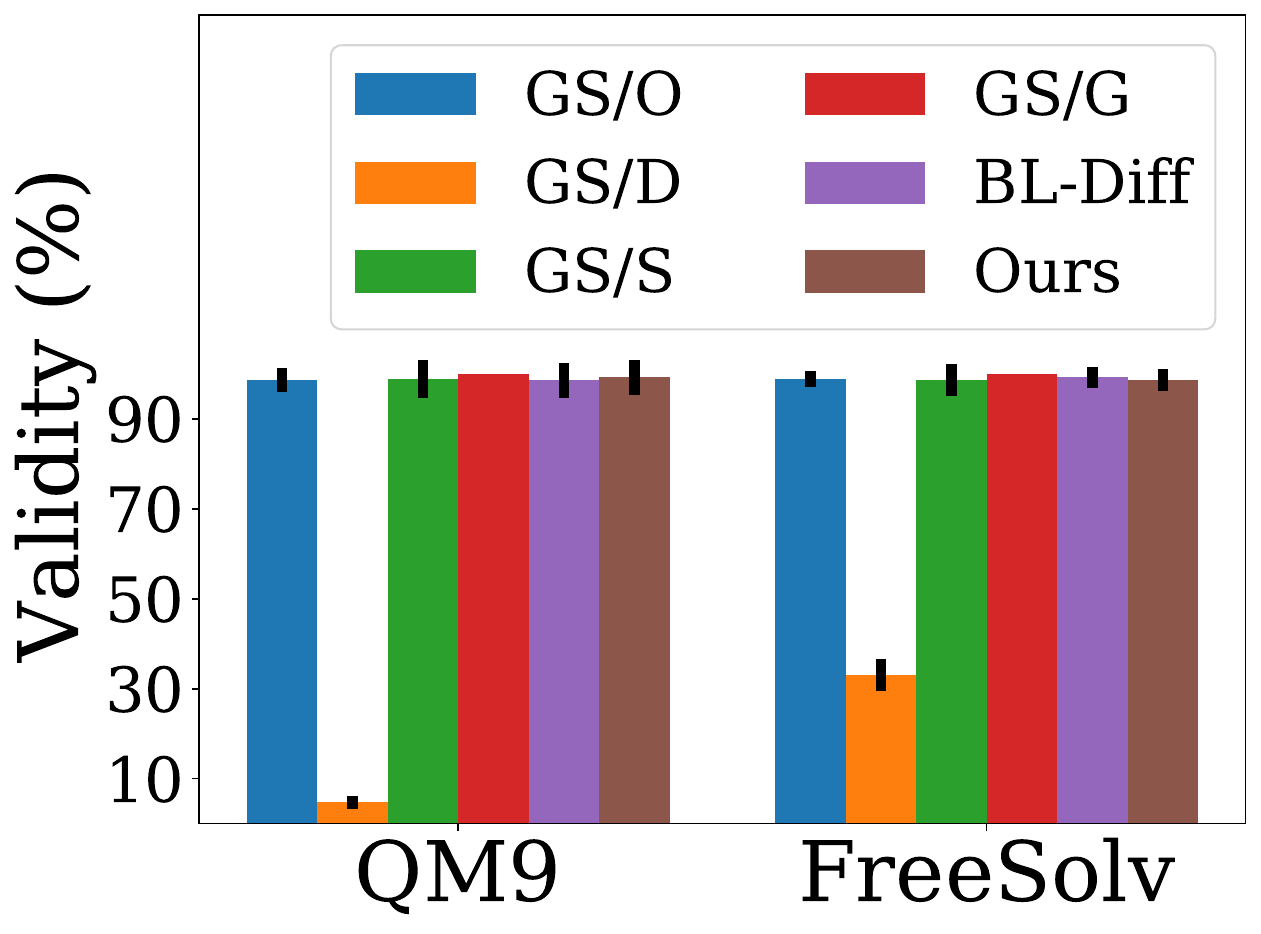}
        % \vskip -0.5em
        \caption{Validity}
    \end{subfigure} ~~
    \begin{subfigure}{0.42\linewidth}
        \includegraphics[width=1\linewidth]{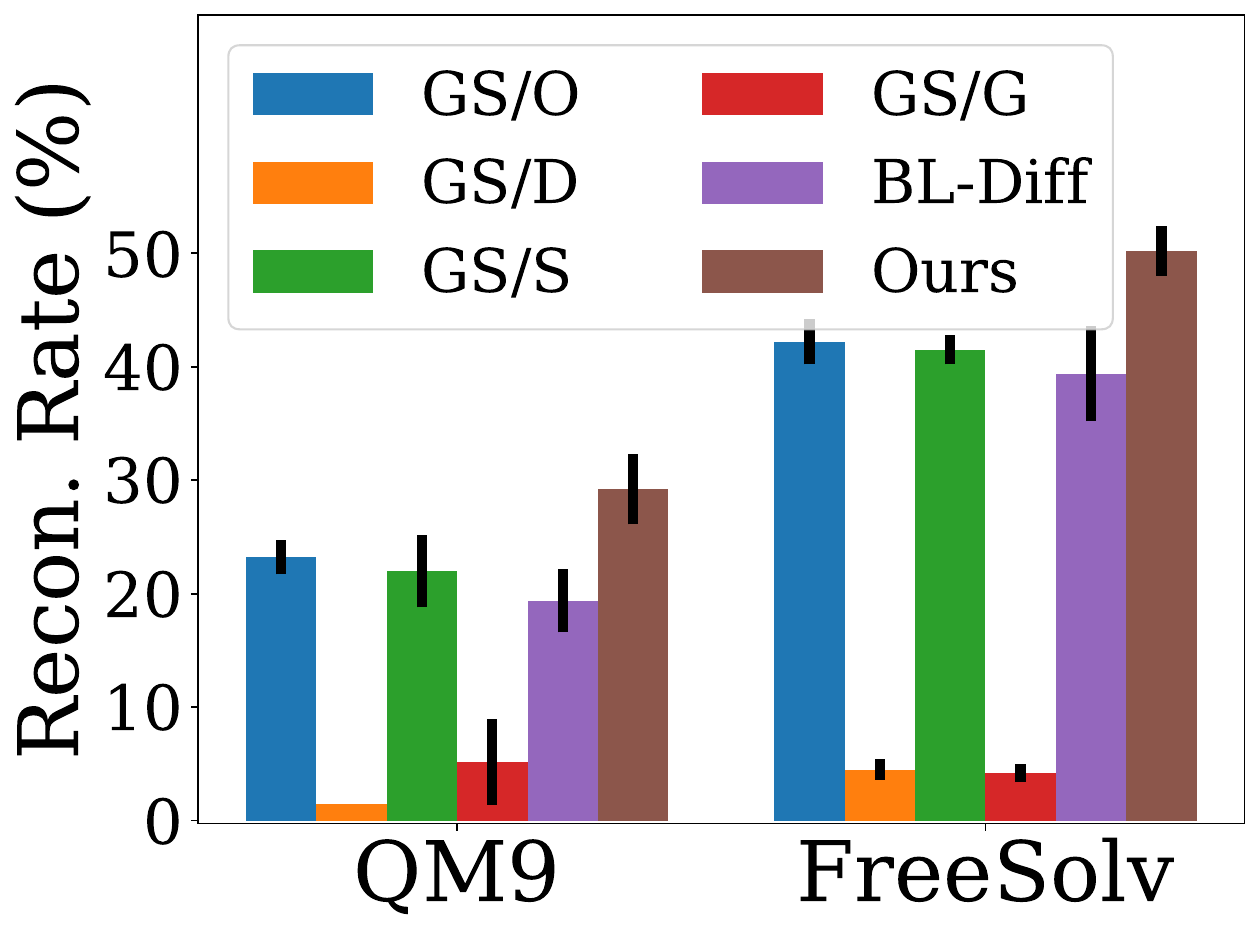}
        % \vskip -0.5em
        \caption{Recon. Rate}
    \end{subfigure}
    % \vskip -1.5em
    \caption{Ablation studies on QM9 and FreeSolv.}
    % \vskip -1em
    \label{fig:ablation_studies_QM9}
\end{figure}

The average results on FreeSolv and QM9 are reported in Fig.~\ref{fig:ablation_studies_QM9}. From the figure, we observe that: (\textbf{i}) GraphSteal/O and GraphSteal/S perform significantly worse than GraphSteal in terms of reconstruction rate, while they still outperform BL-Diff. It shows the effectiveness of our proposed diffusion noise optimization and graph selection mask optimization, respectively; (\textbf{ii}) GraphSteal outperforms GraphSteal/D and GraphSteal/G by a large margin in reconstruction rate. This demonstrates the proposed graph generator can effectively enhance the realism and quality of the reconstructed graphs; and (\textbf{iii}) GraphSteal/D exhibits extremely lower validity than all other methods. This aligns with our description in Sec.~\ref{sec:generating_graphs_via_sdedit} that directly using graphs optimized based on Sec.~\ref{sec:diffusion_noise_optimization} as reconstructions would encounter issues of unrealism and invalidity.

\subsection{Reconstruction Visualization}
\begin{figure}[t]
    \small
    \centering
    \includegraphics[width=0.96\linewidth]{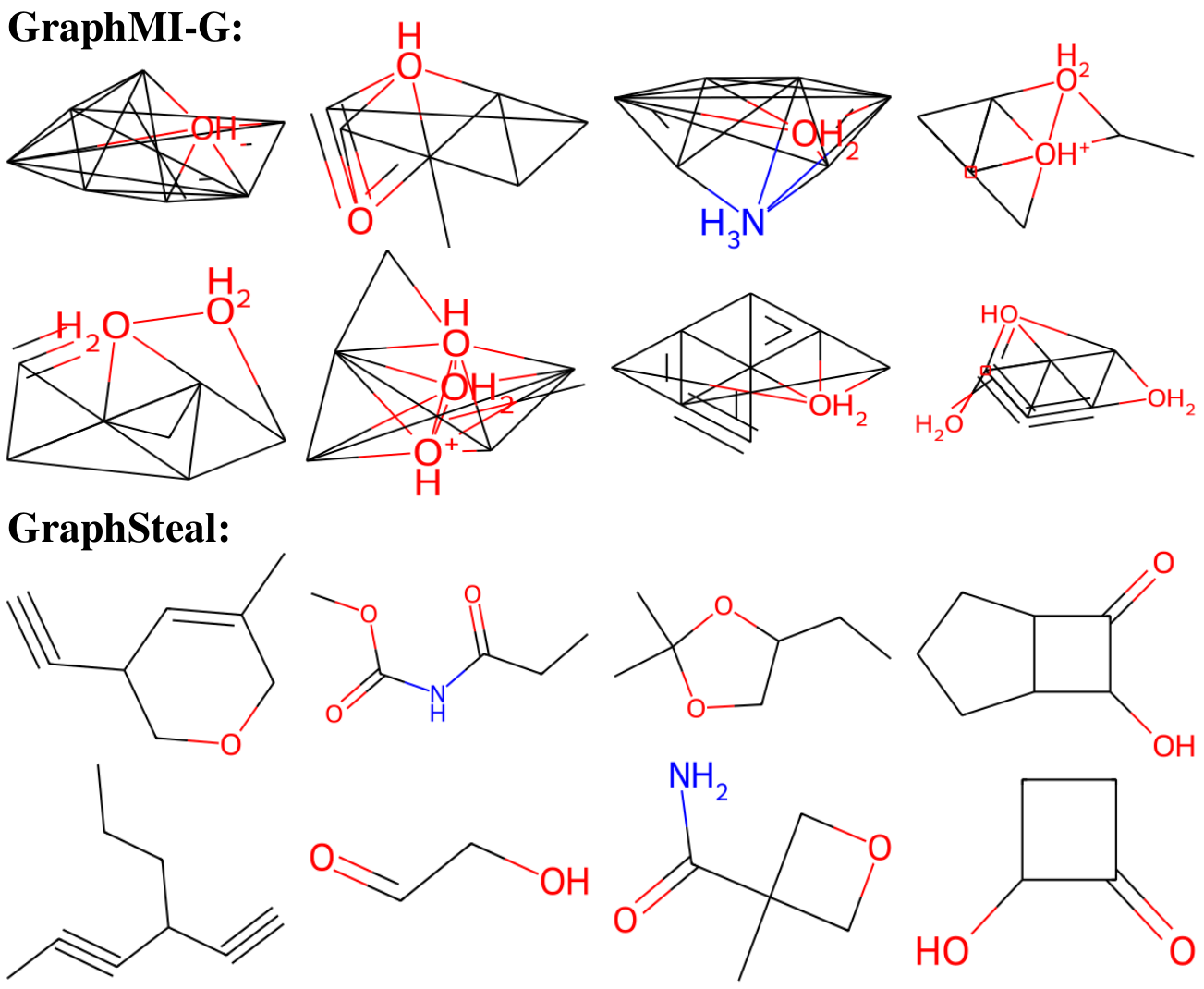}
    % \vskip -1em
    \caption{Reconstructed training graphs of QM9.}
    \label{fig:visualization}
\end{figure}

In this subsection, we conduct a case study to further demonstrate the effectiveness of GraphSteal. We conduct graph stealing attack using both GraphMI-G and GraphSteal on QM9 and then visualize the top-$8$ representative reconstructed graphs. The visualizations are plotted in Fig.~\ref{fig:visualization}. From the figure, we observe that the graphs reconstructed by GraphMI-G appear unrealistic and invalid, which implies the poor performance of GraphMI-G in graph stealing attacks. In contrast, the graphs reconstructed by our method exhibit both high quality and realism, thereby verifying the effectiveness of our method in reconstructing high-quality graphs by leveraging the model parameters of GNNs.

\section{Conclusion and FUTURE WORK}
In this paper, we study a novel privacy attack problem of stealing training graphs from trained GNNs. We propose a novel framework, GraphSteal, to reconstruct training graphs by leveraging the parameters of GNNs. Specifically, we apply a graph diffusion model as the graph generator to reconstruct graphs realistically. We then implement a diffusion noise optimization to enhance the resemblance of the graphs generated by the graph generator to the target training data. Moreover, a model parameter-guided graph selection method is proposed to identify the training graphs from the generated graphs by leveraging the parameters of GNNs.
Extensive experiments on different real-world datasets demonstrate the effectiveness of GraphSteal in reconstructing realistic and high-quality training graphs from trained GNNs. 
There are two directions that need further investigation. First, in this paper, we only focus on performing graph stealing attack in the white-box setting. Thus, it is also interesting to in investigate how to conduct the black-box graph stealing attack effectively. Second, it is also worthwhile to investigate how to defend against the graph stealing attack. The discussions of the potential countermeasures and ethical implications are in \textit{Appendix~\ref{sec:countermeasures}} and \textit{Appendix~\ref{sec:ethical_implications}}, respectively.

\begin{acks}
This material is based upon work supported by, or in part by, the Army Research Office (ARO) under grant number W911NF-21-1-0198, the Department of Homeland Security (DHS) CINA under grant number E205949D, and the Cisco Faculty Research Award. The findings in this paper do not necessarily reflect the view of the funding agencies. 
\end{acks}
% \newpage

% \begin{acks}
% \end{acks}

% \newpage
\bibliographystyle{ACM-Reference-Format}
\bibliography{ref}

\newpage
\appendix
\newpage
\section{Detailed Proofs}
\subsection{Preliminaries of KKT Conditions}
\label{appendix:kkt_points}
In this subsection, we present the backgrounds of the Karush-Kuhn-Tucker (KKT) condition.

Consider the following optimization problem:
\begin{equation}
\begin{aligned}
\label{eq:basic_op_problem_appendix}
    &\min_{\mathbf{x}} {f(\mathbf{x})} \\
    &\text{s.t.} \quad \delta_n(\mathbf{x}) \leq 0 \quad \forall{n\in[N]},
\end{aligned}
\end{equation}
where $f, \delta_1,\ldots,\delta_n:\mathbb{R}^d \rightarrow \mathbb{R}$ are locally Lipschitz functions. We say that $\mathbf{x} \in \mathbb{R}^d$ is a feasible point of Eq.~(\ref{eq:basic_op_problem_appendix}) if $\mathbf{x}$ satisfies $\delta_n(\mathbf{x}) \leq 0$ for all $n \in [N]$. This constrained optimization problem can then form the Lagrangian function: 
\begin{equation}
    L = f(\mathbf{x}) + \sum_{n=1}^{N}\lambda_n\delta_n(\mathbf{x}),
\end{equation}
if there exist $\lambda_n\in\lambda$ satisfies the following conditions:
\begin{align}
    &\nabla_{\mathbf{x}}L = \nabla{f} + \sum_{n=1}^{N}\nabla{\delta_n} = \mathbf{0} \label{eq:basic_stationarity_appendix}\\
    &\delta_n(\mathbf{x}) \leq 0,\\
    &\lambda_1, \ldots,\lambda_N \geq 0,\\
    &\lambda_n \delta_n(\mathbf{x}) = 0, \quad  \forall {i\in[N]}\label{eq:basic_complementary_appendix},
\end{align}
where Eq.~(\ref{eq:basic_stationarity_appendix}) to Eq.~(\ref{eq:basic_complementary_appendix}) are the stationarity, primal feasibility, dual feasibility and complementary slackness of KKT conditions, respectively. 
Then we give the definition of KKT point as follows:
\begin{definition}[KKT point]
    A feasible point $\mathbf{x}$ of the optimization problem in Eq.~(\ref{eq:basic_op_problem_appendix}) is KKT point if $\mathbf{x}$ satisfies KKT conditions in Eq.~(\ref{eq:basic_stationarity_appendix}) to Eq.~(\ref{eq:basic_complementary_appendix}).
\end{definition}
Note that a global minimum of Eq.~(\ref{eq:basic_op_problem_appendix}) may not be a KKT point, but under some regularity assumptions (e.g., Mangasarian-Fromovitz Constraint Qualification), the KKT conditions will then be the necessary condition for global optimality. 

\subsection{Homogeneity of Graph Neural Networks}
\label{sec:homogeneity_GNN_appendix}
Before we start to discuss the homogeneity of GNNs, we recall the definition of homogeneous neural networks in Sec.~\ref{sec:homogeneous_graph_neural_networks} as follows:

\textbf{Definition~\ref{def:homogeneous_neural_network}} (Homogeneous Neural Networks~\cite{Lyu2020Gradient}).
\textit{Let $f$ be a neural network with model parameters as ${\boldsymbol{\theta}}$. Then $f$ is a \textit{homogeneous} neural network if there is a number $L>0$ such that the model output $f(\mathcal{G};{\boldsymbol{\theta}})$ satisfies the following equation:
    \begin{equation}
    \label{eq:homogeneous_neural_networks_appendix}
    f(\mathcal{G};\sigma{\boldsymbol{\theta}})=\sigma^{L}f(\mathcal{G};{\boldsymbol{\theta}}),~~\forall{\sigma>0}.
    \end{equation}}

\begin{table*}[h]
    \centering
    \caption{Averaged results of the absolute number of the reconstructed graphs.}
    \vskip -0.8em
    \resizebox{0.70\textwidth}{!}
    {\begin{tabularx}{0.70\linewidth}{p{0.08\linewidth}CCCCCCc}
    \toprule 
    {Dataset} & {10} & 
    {20} & {50}& {100}& {200}& {500}\\
     \midrule
\multirow{1}{*}{FreeSolv}
& \multirow{1}{*}{5.9} & 11.4  & 27.5 & 50.4 & 92.6 & {216.0}\\
    
\multirow{1}{*}{QM9}
& \multirow{1}{*}{3.8} & 6.7  & 15.6 & 29.2 & 100 & {121.5}\\
    \bottomrule 
        \end{tabularx}}
    % \vskip -3.em
    \vskip -1.em
    \label{tab:absolute_number_reconstructed_graphs}
\end{table*}
Following~\cite{nacson2019lexicographic,Lyu2020Gradient}, any fully-connected or convolutional neural network with ReLU activations is homogeneous w.r.t the mode parameters $\boldsymbol{\theta}$ if it does not have any bias terms or skip-connections. To extend to GNNs, the message passing in GNNs can be regarded as a generalized form of convolution from the spectral perspective~\cite{kipf2016semi}. Therefore, we can claim that any message-passing based GNNs (e.g. GCN~\cite{kipf2016semi} and SGC~\cite{wu2019simplifying}) with ReLU activations is homogeneous w.r.t the parameters ${\boldsymbol{\theta}}$ if it does not have any skip-connections (e.g., GraphSage~\cite{hamilton2017inductive}) or bias terms, except possibly for the first layer. Our experimental results in Sec.~\ref{sec:flexibility_to_model_arch} also demonstrate the effectiveness of GraphSteal across various GNN architectures. 

We further take SGC~\cite{wu2019simplifying} and GCN as examples.

\subsubsection{Proof for SGC} Given a $K$-layer SGC, the model output $f(\mathcal{G};{\boldsymbol{\theta}})$ is:
\begin{equation}
f(\mathcal{G};{\boldsymbol{\theta}}) = \mathbf{S}^{K}\mathbf{X}\boldsymbol{\theta},
\end{equation}
where $\mathbf{S}=\tilde{\mathbf{D}}^{-1/2}\tilde{\mathbf{A}}\tilde{\mathbf{D}}^{-1/2}$, $\tilde{\mathbf{A}} = \mathbf{A} + \mathbf{I}$ and $\tilde{\mathbf{D}}$ is the degree matrix of $\tilde{\mathbf{A}}$. $\boldsymbol{\theta} = \boldsymbol{\theta}^{(1)}\boldsymbol{\theta}^{(2)}\ldots\boldsymbol{\theta}^{(L)}$
Therefore, for $f(\mathcal{G};{\sigma\boldsymbol{\theta}})$, 
we have:
\begin{equation}
\begin{aligned}
    f(\mathcal{G};{\sigma\boldsymbol{\theta}}) & = \sigma\mathbf{S}^{K}\mathbf{X}\boldsymbol{\theta}\\
    & = \sigma f(\mathcal{G};{\boldsymbol{\theta}}).
\end{aligned}
\end{equation}
Thus, $f(\mathcal{G};{\boldsymbol{\theta}})$ satisfies Eq.~(\ref{eq:homogeneous_neural_networks_appendix}) when $L=1$, which shows that SGC is a homogeneous GNN. 

\subsubsection{Proof for GCN} GCN~\cite{kipf2016semi} can also be proved as a homogeneous GNN in a similar way as GCN is the ReLU version of SGC. We first prove the homogeneity of a single layer of GCN. Specifically, for the $i$-th layer of GCN, the output is:
\begin{equation}
\label{eq:i_th_layer_GCN}
f^{(i)}(\mathbf{H}^{(i-1)};{\boldsymbol{\theta}^{(i)}}) = \text{ReLU}(\mathbf{S}\mathbf{H}^{(i-1)}\boldsymbol{\theta}^{(i)}), \quad i\in\{1,\ldots,K\}
\end{equation}
where $\mathbf{H}^{(i-1)} = f^{(i-1)}(\mathbf{H}^{(i-2)};{\boldsymbol{\theta}^{(i-1)}})$ is the output from the $(i-1)$-th layer of a K-layer GCN, and $\mathbf{H}^{(0)}=\mathbf{X}$.

Then, we apply the positive scalar $\sigma > 0$ to $\boldsymbol{\theta}^{(i)}$ and obtain
\begin{equation}
\begin{aligned}
    f^{(i)}(\mathbf{H}^{(i-1)};{\sigma\boldsymbol{\theta}^{(i)}}) & = \text{ReLU}(\mathbf{S}\mathbf{H}^{(i-1)}\cdot\sigma\boldsymbol{\theta}^{(i)})\\
    & = \text{ReLU}(\sigma\cdot\mathbf{S}\mathbf{H}^{(i-1)}\boldsymbol{\theta}^{(i)}).
\end{aligned}
\end{equation}
If $\mathbf{S}\mathbf{H}^{(i-1)}\boldsymbol{\theta}^{(i)} > 0$, we have 
\begin{equation}
\begin{aligned}
   \text{ReLU}(\sigma\cdot\mathbf{S}\mathbf{H}^{(i-1)}\boldsymbol{\theta}^{(i)}) & = \sigma\cdot\mathbf{S}\mathbf{H}^{(i-1)}\boldsymbol{\theta}^{(i)} \\
   & = \sigma\cdot\text{ReLU}(\mathbf{S}\mathbf{H}^{(i-1)}\boldsymbol{\theta}^{(i)}).  
\end{aligned}
\end{equation} 
And if $\mathbf{S}\mathbf{H}^{(i-1)}\boldsymbol{\theta}^{(i)} \le 0$, we have 
\begin{equation}
\begin{aligned}
    \text{ReLU}(\sigma\cdot\mathbf{S}\mathbf{H}^{(i-1)}\boldsymbol{\theta}^{(i)}) = 0
\end{aligned}
\end{equation}
for all $\sigma>0$, which is consistent with $\text{ReLU}(\mathbf{S}\mathbf{H}^{(i-1)}\boldsymbol{\theta}^{(i)}) = \sigma\cdot\text{ReLU}(\mathbf{S}\mathbf{H}^{(i-1)}\boldsymbol{\theta}^{(i)})=0$. 

Combining these two cases, for the $i$-th GCN layer  $f^{(i)}(\mathbf{H}^{(i-1)};{\boldsymbol{\theta}^{(i)}}) \allowbreak= \text{ReLU}(\mathbf{S}\mathbf{H}^{(i-1)}\boldsymbol{\theta}^{(i)})$, we have
\begin{equation}
\begin{aligned}
    f^{(i)}(\mathbf{H}^{(i-1)};{\sigma\boldsymbol{\theta}^{(i)}}) = \sigma\cdot f^{(i)}(\mathbf{H}^{(i-1)};{\sigma\boldsymbol{\theta}^{(i)}}).
\end{aligned}
\end{equation}
Therefore, there exist $L=1$ such that $f^{(i)}(\mathbf{H}^{(i-1)};{\sigma\boldsymbol{\theta}^{(i)}}) = \sigma^{L}f^{(i)}\allowbreak(\mathbf{H}^{(i-1)};{\sigma\boldsymbol{\theta}^{(i)}})$. Hence, $f^{(i)}(\mathcal{G};{\sigma\boldsymbol{\theta}^{(i)}})$ is a homogeneous function. 

With this, we extend the proof to multiple layers of GCN. Since the output of one layer becomes the input to the next layer, based on Eq.~(\ref{eq:i_th_layer_GCN}), for the $(K-1)$-th GCN layer $f^{(K-1)}(\mathcal{G};\theta)$, we have
\begin{equation}
\begin{aligned}
    f^{(K)}(\mathcal{G};{\boldsymbol{\theta}^{(K)}}) &=  \text{ReLU}(\mathbf{S}f^{(K-1)}(\mathcal{G};{\boldsymbol{\theta}^{(K-1)}})\boldsymbol{\theta}^{(K)})\\
    & = \text{ReLU}(\mathbf{S}\mathbf{H}^{(K-1)}\boldsymbol{\theta}^{(K)}).
\end{aligned}
\end{equation}
When the parameter $\theta^{(1)}$ of the first layer GCN is scaled by $\sigma$, we have
\begin{equation}
\label{eq:first_layer_GCN_scale_sigma}
\begin{aligned}
    f^{(1)}(\mathcal{G};{\sigma\boldsymbol{\theta}^{(1)}}) & = \text{ReLU}(\mathbf{S}\mathbf{H}^{(i-1)}\cdot\sigma\boldsymbol{\theta}^{(i)})\\
    & = \sigma f^{(1)}(\mathcal{G};{\boldsymbol{\theta}^{(1)}})\\
    & =\sigma\mathbf{H^{(1)}}
\end{aligned}
\end{equation}
Then, when input Eq.~(\ref{eq:first_layer_GCN_scale_sigma}) to the second layer GCN, we have 
\begin{equation}
\begin{aligned}
    f^{(2)}(\sigma\mathbf{H^{(1)}};{\boldsymbol{\theta}^{(2)}}) & = \text{ReLU}(\mathbf{S}\cdot\sigma\mathbf{H}^{(1)}\boldsymbol{\theta}^{(2)})\\
    & = \text{ReLU}(\mathbf{S}\mathbf{H}^{(1)}\cdot\sigma\boldsymbol{\theta}^{(2)})\\
    & = f^{(2)}(\mathbf{H^{(1)}};{\sigma\boldsymbol{\theta}^{(2)}}) \\
    & = \sigma\cdot\text{ReLU}(\mathbf{S}\mathbf{H}^{(1)}\boldsymbol{\theta}^{(2)})\\
    & = \sigma f^{(2)}(\mathbf{H}^{(1)};{\boldsymbol{\theta}^{(2)}}) \\
    & = \sigma \mathbf{H}^{(2)}.
\end{aligned}
\end{equation}
Similarly, we can easily conclude that the output of the $(K-1)$-th layer is 
\begin{equation}
    f^{(K-1)}(\sigma\mathbf{H^{(K-2)}};{\boldsymbol{\theta}^{(K-1)}}) = \sigma \mathbf{H}^{(K-1)}.
\end{equation}
By using the homogeneity of the single-layer GCN, we can have
\begin{equation}
\begin{aligned}
     f^{(K)}(\mathbf{H^{(K-1)}};\sigma\boldsymbol{\theta}^{(K)}) & = \text{ReLU}
     (\mathbf{S}\cdot\sigma\mathbf{H}^{(K-1)}\boldsymbol{\theta}^{(K)}) \\
     & = \text{ReLU}
     (\mathbf{S}\mathbf{H}^{(K-1)}\cdot\sigma\boldsymbol{\theta}^{(K)}) \\
     & = \sigma\cdot\text{ReLU}
     (\mathbf{S}\mathbf{H}^{(K-1)}\boldsymbol{\theta}^{(K)}) \\
     & = \sigma\cdot f^{(K)}(\mathbf{H^{(K-1)}};{\boldsymbol{\theta}^{(K)}}).
\end{aligned}
\end{equation}

Hence, we can conclude that for the $K$ layer GCN, $f(\mathcal{G};\sigma\boldsymbol{\theta}) = \sigma\cdot{f(\mathcal{G};\boldsymbol{\theta})}$, the homogeneity of the $K$ layer GCN is then proved.

Note that our theorem is also applicable to LeakyReLU activation function. The proof is similar to the above.

\subsection{Proof of Theorem~\ref{thm:relationship_parameter_training_data}}
\label{appendix:proof_of_thm_relationship_parameter_training_data}
Before beginning our proof, we first recall Lemma~\ref{lemma:gradient_descent} in Sec.~\ref{sec:homogeneous_graph_neural_networks}.
\textbf{Lemma~\ref{lemma:gradient_descent}} (\cite{Lyu2020Gradient}).
\textit{Let $f_{{\boldsymbol{\theta}}}$ be a homogeneous ReLU neural network with model parameters ${\boldsymbol{\theta}}$. Let $\mathcal{D}_t=\{(\mathcal{G}_{i},y_{i})\}^{|\mathcal{D}_t|}_{i=1}$ be the classification training dataset of $f_{{\boldsymbol{\theta}}}$, where $y_i\in\{1,\ldots,C\}$. Assume $f_{{\boldsymbol{\theta}}}$ is trained by minimizing the cross entropy loss 
    $\mathcal{L}_{ce} = \sum_{i=1}^{|\mathcal{D}_t|}\log({1+\sum_{j\neq y_i}e^{-q_{ij}(\boldsymbol{\theta})}})$ over $\mathcal{D}_t$ using gradient flow, where $q_{ij}(\boldsymbol{\theta}) = f_{\boldsymbol{\theta}}(\mathcal{G}_i)_{y_i} - f_{\boldsymbol{\theta}}(\mathcal{G}_i)_{j}$ and $f_{\boldsymbol{\theta}}(\mathcal{G}_i)_{y_i}$ denotes the $y_i$-th entry of the logit score vector $f_{\boldsymbol{\theta}}(\mathcal{G}_i)$ before applying the softmax normalization. Then, gradient flow converges in direction towards a first-order stationary point of the following max-margin optimization problem:
    \begin{equation}
        \label{eq:max_margin_optimization_appendix}
        \underset{{{\boldsymbol{\theta}}}}{\min}\frac{1}{2}||{{\boldsymbol{\theta}}}||^2_2, \quad \text{s.t.} \quad 
        q_{ij}(\boldsymbol{\theta}) \geq 1,~~\forall{i\in[|\mathcal{D}_t|]}, {j\in[C]\backslash \{y_i\}}. 
        % f_{{{\boldsymbol{\theta}}}}(\mathcal{G}_i)\cdot{y_i}\geq 1,~~\forall{i\in[|\mathcal{D}_t|]}
    \end{equation}
    }

Given a homogeneous ReLU GNN model $f_{\boldsymbol{\theta}}$, according to Lemma~\ref{lemma:gradient_descent}, the training of $f_{\boldsymbol{\theta}}$ using gradient descent can be viewed as solving a max-margin optimization problem to maximize the margin $q_{ij}(\boldsymbol{\theta}) = f_{\boldsymbol{\theta}}(\mathcal{G}_i)_{y_i} - f_{\boldsymbol{\theta}}(\mathcal{G}_i)_{j}$ over all possible directions. 

Moreover, the optimization problem in Eq.~(\ref{eq:max_margin_optimization_appendix}) can be reformulated to the following constrained optimization problem:
\begin{equation}
    \label{eq:reform_max_margin_optimization_appendix}
    \underset{{{\boldsymbol{\theta}}}}{\min}\frac{1}{2}||{{\boldsymbol{\theta}}}||^2_2, \quad \text{s.t.} \quad 
        \beta_{i}(\boldsymbol{\theta}) \geq 1,~~\forall{i\in[|\mathcal{D}_t|]}, 
\end{equation}
where $\beta_{i}(\boldsymbol{\theta}) = f_{{\boldsymbol{\theta}}}(\mathcal{G}_i)_{y_i} - \max_{j\neq y_i}\{f_{{{\boldsymbol{\theta}}}}(\mathcal{G}_i)_{j}\}$ indicates the margin of $f_{\boldsymbol{\theta}}$ for a single data point $(\mathcal{G}_i,y_i)$. Following from~\cite{Lyu2020Gradient}, it can be proved that Eq.~(\ref{eq:reform_max_margin_optimization_appendix}) satisfies the Mangasarian-Fromovitz Constraint Qualification (MFCQ). KKT conditions are then first-order necessary conditions for the global optimality of Eq.~(\ref{eq:reform_max_margin_optimization_appendix}). Thus, after convergence of training $f_{\boldsymbol{\theta}}$ in direction to a KKT point $\tilde{\boldsymbol{\theta}}$ using gradient descent, there exist $\lambda_1, \dots, \lambda_{|\mathcal{D}_t|}\in\mathbb{R}$ such that:
\begin{align}
    % \label{eq:kkt_stationary}
    &\tilde{{\boldsymbol{\theta}}} = \sum_{i=1}^{|\mathcal{D}_t|}\lambda_{i}(\nabla_{{\boldsymbol{\theta}}}f_{\tilde{{\boldsymbol{\theta}}}}(\mathcal{G}_i)_{y_i} - \nabla_{{\boldsymbol{\theta}}}\max_{j\neq y_i}\{f_{\tilde{{\boldsymbol{\theta}}}}(\mathcal{G}_i)_{j}\}),\label{eq:stationarity_appendix}\\ 
    &\beta_{i}(\tilde{\boldsymbol{\theta}}) = f_{{\tilde{\boldsymbol{\theta}}}}(\mathcal{G}_i)_{y_i} - \max_{j\neq y_i}\{f_{{{\tilde{\boldsymbol{\theta}}}}}(\mathcal{G}_i)_{j}\} \geq 1,~~\forall{i\in[|\mathcal{D}_t|]}\\
    % &f_{\overline{{\boldsymbol{\theta}}}}(\mathcal{G}_i)y_i>1,~~\forall{i\in[|\mathcal{D}_t|]}, \\
    &\lambda_1, \dots, \lambda_{|\mathcal{D}_t|} \geq 0, \\
    & \lambda_i=0 \quad \text{if}~~\beta_{i}(\tilde{\boldsymbol{\theta}})\neq{1}, \forall {i\in |\mathcal{D}_t|}\label{eq:complementary_slackness_appendix},
\end{align}
where 
$\mathcal{G}_i\in\mathcal{D}_t$ denote the graph with ground-truth label $y_{i}$ for training $f_{{\boldsymbol{\theta}}}$. Eq.~(\ref{eq:stationarity_appendix}) to Eq.~(\ref{eq:complementary_slackness_appendix}) are the stationarity, primal feasibility, dual feasibility and complementary slackness of KKT conditions, respectively. Therefore, the KKT point $\tilde{\boldsymbol{\theta}}$ is the point that satisfy Eq.~(\ref{eq:kkt_stationary}) in Theorem~\ref{thm:relationship_parameter_training_data}. This completes our proof.

\begin{figure}[t]
    \small
    \centering
    \begin{subfigure}{0.49\linewidth}
        \includegraphics[width=0.98\linewidth]{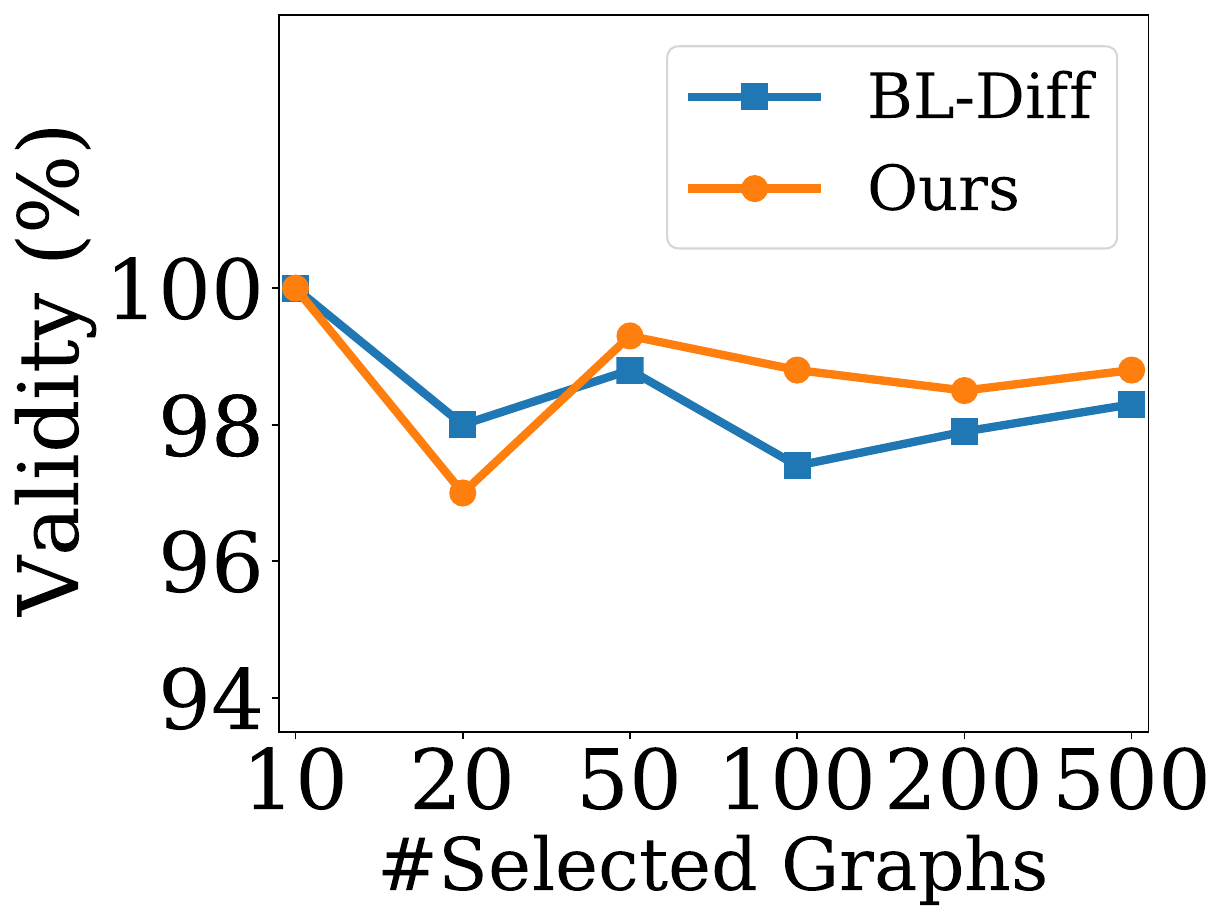}
        \vskip -0.5em
        \caption{Validity}
    \end{subfigure}
    \begin{subfigure}{0.49\linewidth}
        \includegraphics[width=0.98\linewidth]{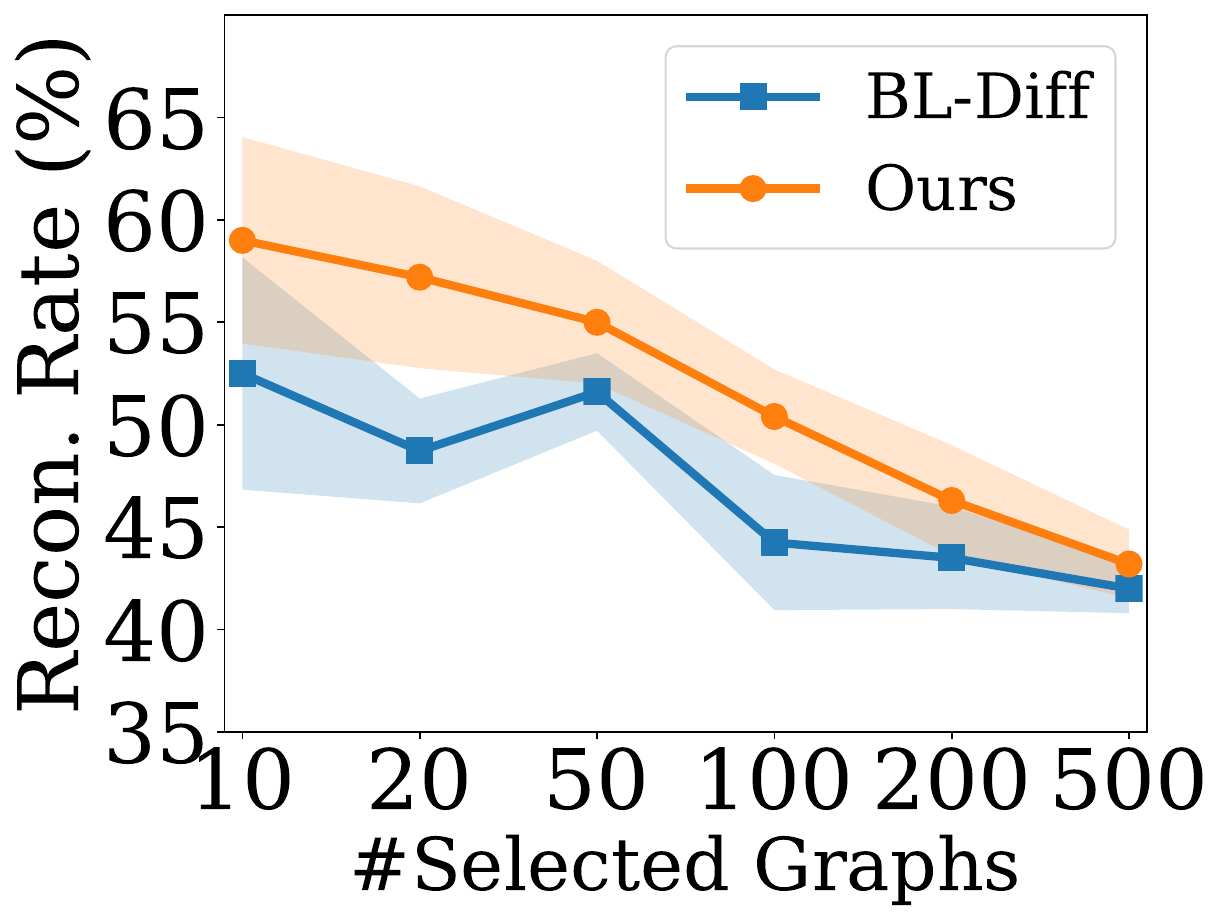}
        \vskip -0.5em
        \caption{Recon. Rate}
    \end{subfigure}
    \vskip -1.5em
    \caption{Impact of the numbers of selected graphs on FreeSolv}
    \vskip -2em
    \label{fig:impact_numbers_of_selected_graphs_Freesolv_appendix}
\end{figure}

\section{Additonal Details of Experiment Settings}
\subsection{Dataset Settings}
We conduct experiments on 3 well-known datasets, i.e., FreeSolv, ESOL and QM9~\cite{wu2018moleculenet}, that are widely used for various graph-level tasks, including graph generation and graph regression. Given our focus on graph classification, we adapt these datasets to our setting by categorizing each graph according to the magnitude of its regression values and ensuring an equal division of graphs within each class. More specifically, for FreeSolv and ESOL, we partition the datasets into two classes, maintaining an equal number of graphs in each. To further validate the performance of GraphSteal in the multi-class graph classification task, we divide the QM9 dataset into three classes, also ensuring each class contains an equal number of graphs.

\label{appendix:datasets}

\subsection{Additional Details of Evaluation Metrics}
\label{appendix:evaluation_metrics}
\subsubsection{Fréchet ChemNet Distance (FCD)} \label{appendix:FCD}
FCD~\cite{preuer2018fcd} is a popular metric used to calculate the distance between the distribution $p_{w}(\cdot)$ of real-world molecules and the distribution $p(\cdot)$ of molecules from a generative model. To calculate this distance, firstly, we use the activations of the penultimate layer of ChemNet~\cite{goli2020chemnet} to obtain the representation of each molecule. We then calculate the mean and covariance of these activations for the two distributions. The two distributions $(p(\cdot), p_{w}(\cdot))$ are compared using the Wasserstein-2 distance~\cite{vaserstein1969markov}. Formally, FCD $d(\cdot,\cdot)$ is given by:
\begin{equation}
    d^2((\mathbf{m},\mathbf{C}),(\mathbf{m}_w,\mathbf{C}_w)) = ||\mathbf{m}-\mathbf{m}_{w}||^2_2 + \text{Tr}(\mathbf{C}+\mathbf{C}_w-2\cdot(\mathbf{C}\mathbf{C}_w)^{1/2}),
\end{equation}
where $(\mathbf{m}_w,\mathbf{C}_w)$ are the mean and covariance of Gaussian $p_w(\cdot)$ obtained from real-world, and $(\mathbf{m},\mathbf{C})$ are the mean and covariance of Gaussian $p(\cdot)$ obtained from the generative model. Note that FCD here is reported as $d^2(\cdot,\cdot)$ analogously to~\cite{heusel2017gans}.

\subsubsection{How To Evaluate the Exactly Graph Matching?}
\label{appendix:additional_details_graph_matching}
We utilize SMILES strings of molecule graphs for graph matching. SMILES (Simplified Molecular Input Line Entry System)~\cite{weininger1988smiles} is a widely used notation in chem-informatics for encoding the structure of chemical molecules using short ASCII strings, thereby facilitating efficient representation and manipulation of chemical structures. The reconstructed graph is matched with the target graph if their SMILES strings are identical~\cite{landrum2013rdkit, o2016comparing}. Specifically, let $\mathcal{T}_r = \{\tau_1,\ldots,\tau_m\}$ and $\mathcal{T}_t = \{\tau_1,\ldots,\tau_m\}$ denote the SMILES strings sets of reconstructed graphs $\mathcal{D}_r$ and training graphs $\mathcal{D}_t$, respectively. Then, $\text{Reconstruction rate}= |\mathcal{T}_r\cap\mathcal{T}_t|/|\mathcal{T}_r|$, where $|\mathcal{T}_r\cap\mathcal{T}_t|$ denotes the number of the SMILES strings of reconstructed graphs exactly matched with those of graphs in $\mathcal{D}_t$.
This method is efficient and reasonable for matching molecule graphs as SMILES strings cover various aspects of molecular structure, encoding information about atoms, bonds, and connectivity within a molecule.
% \section{Additional Results of Reconstruction}
\section{Additional Results of the Impact of Selected Graphs}
\label{appendix:additional_impact_selected_graphs}
We report the validity and reconstruction rate on FreeSolv in Fig.~\ref{fig:impact_numbers_of_selected_graphs_Freesolv_appendix}. Moreover, we also report the absolute number of the reconstructed graphs of GraphSteal on FreeSolv and QM9 in Tab.~\ref{tab:absolute_number_reconstructed_graphs}. The observations are similar to those of Fig.~\ref{fig:impact_numbers_of_selected_graphs} in Sec.~\ref{sec:impact_of_the_number_of_selected_graphs}.

\begin{table*}[t]
    \centering
    % \scriptsize
    %\vskip -1em
    \caption{Reconstruction results of GraphSteal against GCN trained with differential privacy on QM9 dataset.}
    \vskip -0.8em
    \resizebox{0.6\textwidth}{!}
    {\begin{tabularx}{0.6\linewidth}{p{0.12\linewidth}CCCc}
    \toprule 
    {Metrics} & {$\epsilon=1.0$} & 
    {$\epsilon=5.0$} & {$\epsilon=10.0$}& {no DP}\\
     \midrule
Validity (\%) $\uparrow$
& 98.6$\pm$1.1 & 99.2$\pm$1.1  & 98.8$\pm$1.3 & 98.8$\pm$0.8 \\

Uniqueness (\%) $\uparrow$
& 100$\pm$0 & 100$\pm$0  & 100$\pm$0 & 100$\pm$0 \\

Recon. Rate (\%) $\uparrow$
& 22.4$\pm$2.4 & 24.0$\pm$3.6  & 26.0$\pm$0.7 & 29.2$\pm$3.1 \\

FCD (\%) $\downarrow$
& 2.3$\pm$0.1 & 2.3$\pm$0.2  & 2.1$\pm$0.2 & 2.0$\pm$0.2 \\
    \bottomrule 
        \end{tabularx}}
    % \vskip -3.em
    \vskip -1.em
    \label{tab:countermeasure}
\end{table*}

\section{Impact of the Training/Auxiliary Set }
\label{sec:impact_ta}
\subsection{Training/Auxiliary Set Distribution Shift }
\label{sec:impact_ta_distribution}
To investigate performance where there is a distribution shift between the auxiliary and target training datasets, we propose to divide QM9 in the following way to make the auxiliary and target training datasets have a distribution shift. Specifically, we first train a GCN with graph labels and use the trained GCN to obtain graph representations. We then apply K-Means to cluster these graphs into 8 groups, setting 1 group as the training set and the remaining groups as the auxiliary set. Hence, there is a significant distribution shift between the two sets. The results on the QM9 dataset are reported in Tab.~\ref{tab:QM9_distribution_shift}. From the table, we can observe that GraphSteal still outperforms the baselines in this setting, demonstrating its effectiveness in the case of there is a significant distribution shift between the training and auxiliary datasets.
\begin{table}[h]
    \centering
    % \scriptsize
    %\vskip -1em
    \caption{Reconstruction results on QM9 for GCN when training and auxiliary sets have a significant distribution shift.}
    \vskip -0.8em
    \resizebox{0.98\linewidth}{!}
    {\begin{tabularx}{1\linewidth}{p{0.18\linewidth}CCCc}
    \toprule 
    {Metrics} & {Valid. (\%) } & 
    {Unique. (\%)} & {Recon. (\%) }& {FCD (\%) }\\
     \midrule
BL-Rand
& \multirow{1}{*}{3.2$\pm$1.1} & \textbf{100$\pm$0}  & 0$\pm$0 & 18.8$\pm$0.8 \\

BL-Diff
& \multirow{1}{*}{\textbf{100$\pm$0}} & 99.6$\pm$0.5  & 1.8$\pm$0.8 & 9.0$\pm$2.4 \\

GraphMI-G
& \multirow{1}{*}{10.0$\pm$9} & \textbf{100$\pm$0}  & 0$\pm$0 & 11.5$\pm$0.5 \\

Ours
& \multirow{1}{*}{\textbf{100$\pm$0}} & \textbf{100$\pm$0 } & \textbf{7.0$\pm$1.8} & \textbf{6.2$\pm$0.3} \\
    \bottomrule 
        \end{tabularx}}
    % \vskip -3.em
    \vskip -1.em
    \label{tab:QM9_distribution_shift}
\end{table}

\subsection{Training/Auxiliary Set Split Ratio}
\label{sec:impact_ta_set_ratio}
In this section, we investigate the impact of the training/auxiliary dataset split (T/A split). We set the T/A split at $\{$10\%/70\%, 50\%/30\%, 60\%/20\%$\}$.  Specifically, 50\%/30\% means that 50\% and 30\% of the original dataset serve as the training set for the GNN classifier and the auxiliary set, respectively. The target GNN model is set as GCN.
All other settings are the same as that in Sec.~\ref{sec:impact_of_the_number_of_selected_graphs}. The results on QM9 are reported in Tab.~\ref{tab:impact_TA_split_QM9_GCN}. From the table, we observe that:
\begin{itemize}[leftmargin=*]
    \item GraphSteal achieves good performance across all T/A split settings. This demonstrates the effectiveness of GraphSteal in various T/A split ratios, especially when the size of the auxiliary set is small.
    \item As the T/A split ratio increases, the reconstruction rate rises. We analyze that the reason is that although the size of the auxiliary dataset decreases as the T/A split ratio increases, it remains sufficient for the graph diffusion model to acquire enough knowledge to generate novel and valid graphs for the reconstruction selection to identify the generated graphs that belong to the target training graphs. Moreover, as the size of the training dataset also increases, our framework can better match the reconstructed graphs with more training graphs, leading to an increase in the reconstruction rate.
\end{itemize}
\begin{table}[t]
    \centering
    % \scriptsize
    %\vskip -1em
    \caption{Impact of the T/A split on QM9 dataset.}
    \vskip -0.8em
    \resizebox{0.9\linewidth}{!}
    {\begin{tabularx}{0.9\linewidth}{p{0.2\linewidth}CCc}
    \toprule 
    {T/A split} & {10\%/70\%} & 
    {50\%/30\%} & {60\%/20\%}\\
     \midrule
Validity (\%) $\uparrow$
& \multirow{1}{*}{98.8$\pm$0.8} & {99.0$\pm$0.7}  & 99.4$\pm$0.5 \\

Unique. (\%) $\uparrow$
& \multirow{1}{*}{{100$\pm$0}} & 98.8$\pm$0.4  & 100$\pm$0  \\

Recon. (\%) $\uparrow$
& \multirow{1}{*}{29.2$\pm$3.1} & {45.0$\pm$1.8}  & 60.6$\pm$4.0  \\

FCD (\%) $\downarrow$
& \multirow{1}{*}{{2.0$\pm$0.2}} & {1.8$\pm$0.3 } & {1.9$\pm$0.3}  \\
    \bottomrule 
        \end{tabularx}}
    % \vskip -3.em
    \vskip -1.em
    \label{tab:impact_TA_split_QM9_GCN}
\end{table}

We further fix the training dataset at 10\% of the original dataset and vary the size of the auxiliary dataset to \{20\%, 30\%, 70\%\} of the original dataset. The results on QM9 are reported in Tab.~\ref{tab:impact_auxiliary_QM9_GCN}. From the table, we observe that GraphSteal consistently achieves good performance across all auxiliary dataset sizes, further validating the effectiveness of GraphSteal in various auxiliary dataset sizes.
\begin{table}[h]
    \centering
    % \scriptsize
    %\vskip -1em
    \caption{Impact of the size of the auxiliary dataset on QM9 dataset.}
    \label{tab:impact_auxiliary_QM9_GCN}
    \vskip -0.8em
    \resizebox{0.9\linewidth}{!}
    {\begin{tabularx}{0.9\linewidth}{p{0.22\linewidth}CCc}
    \toprule 
    {Auxiliary size} & {20\%} & 
    {30\%} & {70\%}\\
     \midrule
Validity (\%) $\uparrow$
& \multirow{1}{*}{99.2$\pm$0.7} & {99.2$\pm$0.8}  & 98.8$\pm$0.8 \\

Unique. (\%) $\uparrow$
& \multirow{1}{*}{{100$\pm$0}} & 100$\pm$0  & 100$\pm$0  \\

Recon. (\%) $\uparrow$
& \multirow{1}{*}{23.2$\pm$2.0} & {25.1$\pm$2.3}  & 29.2$\pm$3.1  \\

FCD (\%) $\downarrow$
& \multirow{1}{*}{{1.8$\pm$0.2}} & {1.7$\pm$0.3} & {2.0$\pm$0.2}  \\
    \bottomrule 
        \end{tabularx}}
    % \vskip -3.em
    \vskip -1.em
\end{table}

\section{Potential Countermeasures}
\label{sec:countermeasures}
In this section, we explore the potential countermeasures of graph stealing attacks. Since graph stealing attacks represent a new type of privacy attack that steals private training graphs from trained GNN classifiers, there is no existing work studying the defense against this attack. One potential defense could be differential privacy (DP). Here, we investigate the effectiveness of DP against graph stealing attacks. Following \cite{abadi2016deep}, we add Gaussian noise to the gradients in each training iteration of target GNN classifier training to ensure $(\epsilon,\delta)-\text{DP}$. We fix $\delta = 10^{-5}$ and vary the noise scale to $\{1.0,5.0,10.0\}$. The reconstruction results on QM9 are reported in Tab.~\ref{tab:countermeasure}. From the table, we can observe that as the privacy budget $\epsilon$ decreases, GraphSteal consistently shows good validity, uniqueness, and FCD but only a slight drop in the reconstruction rate. This indicates that applying differential privacy to GNNs cannot prevent graph stealing attacks effectively. Therefore, there is an emerging need to design more effective countermeasures against GraphSteal.
% {Though one can increase the privacy budget to better protect privacy, the added noise will significantly reduce the accuracy of the GCN classifier. The utility-privacy trade-off of differential privacy makes it less applicable in the real world. Therefore, there is an emerging need to design more effective countermeasures against GraphSteal.} %This indicates that applying DP to GNNs cannot prevent graph stealing attacks, highlighting the need for designing more effective countermeasures against GraphSteal.

\section{Ethical Implications}
\label{sec:ethical_implications}
In this paper, we study a novel privacy attack problem of extracting private training graphs from the trained GNN. Our work uncovers the vulnerability of GNNs to the graph stealing attack and discusses potential countermeasures against this attack. Our work aims to raise awareness about the privacy issues inherent in GNNs and inspire the following works to develop more advanced privacy-preserving methods to protect against graph stealing attacks.  All datasets we used in this paper are publicly available, no sensitive or private dataset from individuals or organizations was used. Our work is mainly for research purposes and complies with ethical standards. Therefore, it does not have any negative ethical impact on society.

\end{document}